# Cardiac MRI-informed electrocardiography enables biologically grounded cardiomyopathy detection

Zhengyao Ding[1*], Kewen Wang[2*], Yujian Hu[3*], Ziyu Li[1], Yiheng Mao[1], Yue Wang[4], Haitao Li[1], Dengnan Zhang[5], Xuanhong Liu[5], Yue Wang[6], Jing Wang[7], Yue Chen[7], Longbo Wang[7], Mengjia Chen[8], Chengchen Zhao[9], Xuesen Chu[10], Yubin Zhang[11], Mor Peleg[12], Judith Somekh[12], Mingxuan Yuan[13], Hui Yan[7], Yue Gu[14, 15], Fan Yi[16], Zhilong Jia[17], Zhentao Tang[13, #], Xi Hu[18], Fei Wu[1, 5], Hongkun Zhang[3], Ting Chen[7, #], Zhengxing Huang[1, 5, #]

[1] College of Computer Science and Technology, Zhejiang University, Hangzhou, China
[2] College of Pharmaceutical Sciences, Zhejiang University, Hangzhou, China
[3] Department of Vascular Surgery, The First Affiliated Hospital of Zhejiang University School of Medicine, Hangzhou, China
[4] Department of Cardiology, Shanghai Ninth People's Hospital, Shanghai Jiaotong University School of Medicine, Shanghai, China
[5] College of Artificial Intelligence, Zhejiang University, Hangzhou, Zhejiang Province, China
[6] Zhejiang University School of Medicine, Hangzhou, China
[7] Department of Cardiology, The First Affiliated Hospital, Zhejiang University School of Medicine, Hangzhou, Zhejiang Province, China
[8] Department of Cardiology, Huzhou Central Hospital, Zhejiang, China
[9] Department of Cardiology, The Second Affiliated Hospital of Zhejiang University School of Medicine, Hangzhou, China
[10] China Ship Scientific Research Center, Wuxi, China
[11] Department of Electrocardiogram, The First Affiliated Hospital, Zhejiang University School of Medicine, Hangzhou, China
[12] Department of Information Systems, University of Haifa, Haifa, Israel
[13] Noah's Ark Lab, Huawei, Beijing, China
[14] Department of Nephrology, Henan Provincial People's Hospital, Zhengzhou, Henan Province, China
[15] Fuwai Central China Cardiovascular Hospital, China
[16] College of Geoinformation Science, Zhejiang University of Technology, Hangzhou, Zhejiang, China
[17] Medical Innovation Research Division of Chinese PLA General Hospital, Beijing, China
[18] Department of Radiology, Sir Run Run Shaw Hospital, Zhejiang University School of Medicine, Hangzhou, Zhejiang Province, China
# Corresponding authors
* These authors contributed equally to this work.

# Abstract

### Background

Cardiac magnetic resonance (CMR) enables detailed myocardial assessment but has limited accessibility. We investigated whether phenotype-guided contrastive learning could enable biologically grounded cardiomyopathy detection using routine electrocardiograms (ECGs).

### Methods

In this retrospective multicentre study, MyoECG was pretrained using 45 241 paired ECG–CMR examinations from UK Biobank and a cardiovascular centre, with 27 quantitative CMR phenotypes guiding representation learning. Development and evaluation involved 1,801,712 ECGs from seven clinical and research sources. We assessed ECG-only cardiomyopathy detection and subtypes classification, benchmarked against five pretrained ECG models, evaluated AI assistance in six cardiologists interpreting 403 ECGs in a reader study, and examined genetic and proteomic associations in UK Biobank.

### Findings

Across cohorts with cardiomyopathy prevalence of 0.2–54.7%, MyoECG achieved areas under the receiver operating characteristic curve ranging from 0.800 (95% CI 0.766–0.830) to 0.974 (0.970–0.978), outperforming all five comparators. In the reader study, mean sensitivity increased by 0.109 (95% CI 0.041–0.175; p=0.009) and specificity by 0.056 (0.004–0.108; p=0.040) with AI assistance. In 41,128 participants, the MyoECG score showed genetic correlations with cardiomyopathy ($r_g$=0.409) and dilated cardiomyopathy ($r_g$=0.371), and associations at four genomic loci. Sixty-nine circulating proteins were associated with the score. After adjustment for CMR phenotypes, genetic correlation with cardiomyopathy attenuated to 0.131 (p=0.143), indicating that the MyoECG score carries CMR-derived structural and functional information rather than dataset-specific shortcuts.

### Interpretation

CMR-informed learning improved ECG-based cardiomyopathy detection, with complementary genetic and proteomic evidence supporting biological plausibility. Prospective studies should establish clinical benefit and downstream imaging burden before implementation.

## Research in Context

### Evidence before this study

We searched PubMed from database inception to Nov 17, 2026, without language restrictions, using "electrocardiogram", "electrocardiography", or "ECG", combined with "artificial intelligence", "deep learning", "representation learning", "cross-modal", or "foundation model", and "cardiomyopathy", "cardiac magnetic resonance", or "cardiac MRI". We focused on original studies of AI-enabled cardiomyopathy detection or ECG–cardiac magnetic resonance (CMR) representation learning and examined relevant reference lists. Previous studies have shown that AI can identify cardiomyopathy and structural cardiac abnormalities from routine ECGs. Cross-modal ECG–CMR approaches have also enabled cardiac phenotype prediction, image synthesis. However, validation beyond single-centre or homogeneous cohorts, particularly across millions of ECGs, remains limited. Performance across populations with differing cardiomyopathy prevalences is also insufficiently characterised. Moreover, to our knowledge, the biological relevance of ECG–CMR information transfer has not been adequately assessed through complementary genetic and proteomic analyses.

### Added value of this study

In this retrospective multicentre study, we developed MyoECG, which uses 27 quantitative CMR phenotypes to guide ECG–CMR learning for ECG-only cardiomyopathy detection and subtype classification. Pretrained on 45,241 paired examinations, MyoECG was developed and evaluated using more than 1.8 million ECGs from seven data sources spanning varying cardiomyopathy prevalences. It outperformed five pretrained ECG models for cardiomyopathy detection, with a six-cardiologist reader study showing improved detection with AI assistance. Genetic and proteomic analyses linked the MyoECG score to cardiomyopathy-related genetic variation and 69 circulating proteins. Attenuation of genetic correlation after adjustment for CMR traits further supported shared information with imaging phenotypes, providing complementary biological evidence for phenotype-guided cross-modal learning.

### Implications of all available evidence

The available evidence suggests that CMR-informed ECG analysis could support opportunistic cardiomyopathy detection and prioritisation of confirmatory imaging, particularly where imaging access is limited. Genetic and proteomic associations provide complementary support for biological plausibility. Prospective multicentre studies should assess diagnostic yield, false-positive burden, calibration, resource use, and patient outcomes before routine implementation.

## Introduction

Cardiomyopathies are heterogeneous myocardial disorders characterized by structural and/or functional abnormalities of the myocardium[1]. Dilated cardiomyopathy (DCM), hypertrophic cardiomyopathy (HCM) and restrictive cardiomyopathy (RCM) are associated with substantial risks of heart failure, arrhythmias and sudden cardiac death, yet affected individuals may remain undiagnosed until symptomatic disease or an index clinical event[2]. Earlier detection can facilitate timely initiation of disease-modifying therapies, evaluation for device implantation and cascade screening of at-risk relatives, thereby providing opportunities to improve long-term clinical outcomes[3]. Current diagnosis relies on the integrated assessment of clinical presentation, echocardiography[4,5], cardiac magnetic resonance imaging (CMR)[6] and, when indicated, genetic testing[7]. Among imaging modalities, CMR provides the most comprehensive non-invasive assessment of ventricular structure, function and myocardial tissue phenotypes[8,9], and plays a central role in cardiomyopathy phenotyping[10]. However, its high cost, limited accessibility, relatively low throughput and requirement for specialized acquisition and interpretation preclude its use for population-scale screening[11]. By contrast, the 12-lead electrocardiogram (ECG) is inexpensive, rapid, non-invasive and widely available in routine clinical practice[12]. However, electrophysiological changes associated with early or subclinical cardiomyopathy are often subtle, spatially distributed and non-specific, making them difficult to capture reliably through visual interpretation or rule-based criteria. Consequently, enabling accurate detection and subtype classification of cardiomyopathy from routine ECGs, without increasing clinical burden, remains a major challenge for scalable cardiovascular screening and diagnostic triage.

Recent advances in artificial intelligence (AI) have substantially expanded the diagnostic capabilities of routine ECG, enabling the inference latent cardiac structure, function and disease risk from routine electrical signals alone[13-15]. Despite these advances, the clinical translation of ECG-AI models for cardiomyopathy remains constrained by several fundamental limitations. First, most existing models are developed and validated in single-centre or otherwise homogeneous cohorts[16], leaving their robustness across healthcare systems, patient populations, ECG devices and acquisition protocols insufficiently tested. Second, prior studies often report performance in balanced or disease-enriched case–control datasets[17], whereas cardiomyopathy prevalence is much uncommon in real-world practice. Consequently, diagnostic metrics reported under artificial prevalence conditions may overestimate clinical utility, particularly because positive and negative predictive values are inherently prevalence dependent. Third, although deep-learning models can achieve high predictive accuracy, they generally provide limited insight into the biological basis of their predictions, making it difficult to distinguish genuine disease-related representations from dataset-specific statistical associations and thereby limiting clinical trust[18]. Collectively, these limitations highlight the need for ECG-AI systems that generalize across diverse clinical settings, maintain performance under real-world disease prevalence, support clinical decision-making, and demonstrate biological validity through concordant genetic, molecular and imaging evidence.

Here we introduce MyoECG (Fig. 1), a CMR-informed and biologically grounded ECG-AI framework for cardiomyopathy detection. Using 45,241 paired ECG–CMR datapoints from UK Biobank (UKB) and Fuwai Central China Cardiovascular Hospital, MyoECG learns ECG representations guided by CMR-derived semantic and phenotypic supervision, and is subsequently fine-tuned to detect cardiomyopathy and classify DCM, HCM and RCM from routine 12-lead ECGs alone. We validate MyoECG across more than 1.8 million ECGs from seven clinical and public datasets spanning a wide range of cardiomyopathy prevalence from 0.2% to 54.7%, assess its incremental diagnostic value for cardiologists with different levels of clinical experience in a reader study, and examine the biological insight of model outputs through genome-wide association analysis, genetic correlation, genetic co-localization, Mendelian randomization, proteomic association and phenome-wide association analyses. Overall, MyoECG provides a scalable approach for CMR-informed ECG features and establishes a framework for linking electrical signals, imaging phenotypes and cardiomyopathy biology.

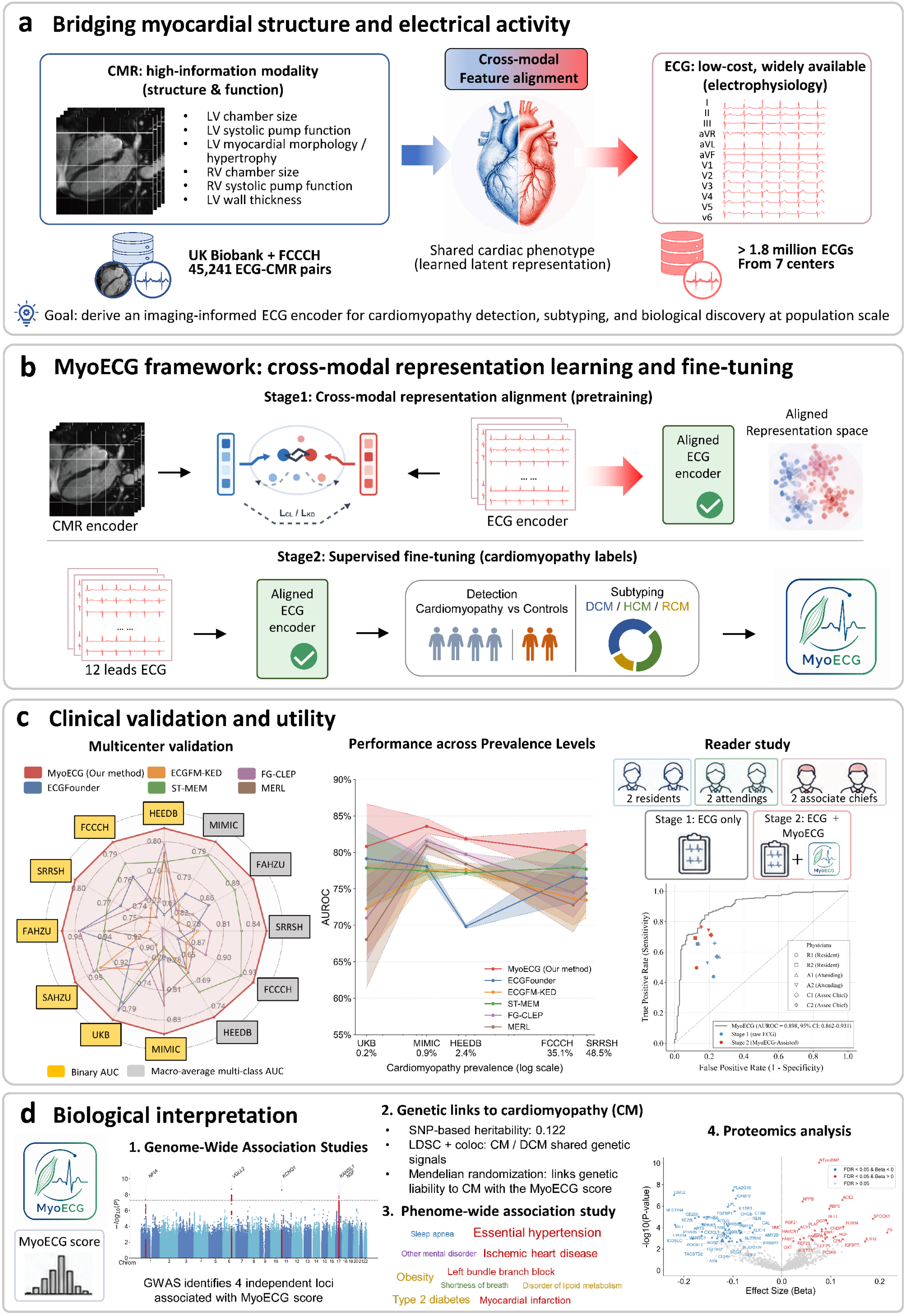

**Fig. 1 | Overview of the MyoECG framework for CMR-informed ECG representation learning, clinical and biological interpretation. a,** Bridging myocardial structure and electrical activity. MyoECG bridges CMR-derived myocardial structural and functional information with routine ECG electrophysiology through

cross-modal feature alignment. Paired ECG–CMR data from UKB and FCCCH were used for multimodal pretraining, and more than 1.8 million ECGs from seven centres or public datasets supported model development and validation. **b,** Cross-modal representation learning and supervised fine-tuning. During pretraining, a CMR encoder provided cardiac semantic representations to guide the ECG encoder, producing a CMR-aligned ECG representation space. In the supervised fine-tuning stage, the aligned ECG encoder was trained for cardiomyopathy detection and subtype classification of dilated cardiomyopathy (DCM), hypertrophic cardiomyopathy (HCM) and restrictive cardiomyopathy (RCM). This ECG-only framework was termed MyoECG. CMR was used only during pretraining; fine-tuning and inference required ECGs alone. **c,** Clinical validation and utility. MyoECG was evaluated for cardiomyopathy detection and subtypes classification across internal, external hospital and public datasets, and was compared with representative pretrained ECG models, including ECGFounder, ECGFM-KED, ST-MEM, FG-CLEP and MERL. Model performance was assessed across datasets with different cardiomyopathy prevalence. A reader study further evaluated whether MyoECG improved ECG interpretation by cardiologists with different experience levels, including residents, attending physicians and associate chief physicians, using a two-stage ECG-only and MyoECG-assisted design. **d,** Biological interpretation. Continuous MyoECG scores were generated using five-fold cross-fitting. Genome-wide association analysis, SNP-based heritability estimation, genetic correlation, colocalization, Mendelian randomization, phenome-wide association analysis and proteomic association analyses were performed to characterize associations between the MyoECG score and cardiomyopathy-related genetic, molecular and phenotypic traits. These analyses provided complementary evidence regarding the biological correlates of the score.

**Table 1 | Characteristics of datasets.** Label provenance differed by dataset and is detailed in Methods. Cardiomyopathy included all cardiomyopathy diagnoses for binary detection; subtype classification was restricted to cardiomyopathy-positive patients with definitive dilated cardiomyopathy (DCM), hypertrophic cardiomyopathy (HCM) or restrictive cardiomyopathy (RCM) labels. Cardiomyopathy subtype percentages were calculated among cardiomyopathy-positive patients, and other cardiovascular disease percentages among non-cardiomyopathy patients. Other cardiovascular disease categories were not mutually exclusive. FAHZU, SAHZU and FCCCH-RS were case-control datasets with relatively healthy controls so comorbid cardiovascular disease prevalence among controls was not reported. Each patient contributed one ECG.

| **Characteristic** | | **UKB** | **FCCCH** | **HEEDB** | **MIMIC** | **SRRSH** | **FAHZU** | **SAHZU** | **FCCCH-RS** |
|---|---|---|---|---|---|---|---|---|---|
| **Number of patients / ECGs** | | 41,128 | 4,113 | 1,588,683 | 159,512 | 1,594 | 5,118 | 464 | 403 |
| **Age (years), mean±s.d.** | | 64.39± 7.81 | 50.45± 16.15 | 54.16± 20.19 | 56.96± 19.36 | 54.46± 15.18 | 61.47 ± 13.88 | 59.76 ± 12.96 | 50.47 ± 14.29 |
| **Sex** | Female | 21,293 (51.8%) | 1,374 (33.4%) | 813,834 (51.2%) | 82,834 (51.9%) | 506 (31.7%) | 1,521 (29.7%) | 203 (43.8%) | 118 (29.3%) |
| | Male | 19,835 (48.2%) | 2,739 (66.6%) | 774,040 (48.7%) | 74,958 (47.0%) | 1,088 (68.3%) | 3,547 (69.3%) | 261 (56.2%) | 285 (70.7%) |
| | Unknown | 0 (0.0%) | 0 (0.0%) | 809 (0.1%) | 1,720 (1.1%) | 0 (0.0%) | 50 (1.0%) | 0 (0.0%) | 0 (0.0%) |
| **Cardiomyopathy status, n (% of dataset)** | Cardiomyopathy | 79 (0.2%) | 1,444 (35.1%) | 37,584 (2.4%) | 1,497 (0.9%) | 773 (48.5%) | 2,741 (53.6%) | 254 (54.7%) | 260 (64.5%) |
| | Non-Cardiomyopathy | 41,049 (99.8%) | 2,669 (64.9%) | 1,551,099 (97.6%) | 158,015 (99.1%) | 821 (51.5%) | 2,377 (46.4%) | 210 (45.3%) | 143 (35.5%) |
| **Cardiomyopathy subtype, n (% of cardiomyopathy patients)** | Dilated cardiomyopathy (DCM) | 26 (32.9%) | 726 (50.3%) | 5,183 (13.8%) | 81 (5.4%) | 270 (34.9%) | 1,681 (61.3%) | NA | 119 (45.8%) |
| | Hypertrophic cardiomyopathy (HCM) | 22 (27.8%) | 431 (29.8%) | 4,214 (11.2%) | 93 (6.2%) | 307 (39.7%) | 900 (32.8%) | | 78 (30.0%) |
| | Restrictive cardiomyopathy (RCM) | 0 (0.0%) | 47 (3.3%) | 365 (1.0%) | 6 (0.4%) | 22 (2.8%) | 160 (5.8%) | | 28 (10.8%) |
| **Other cardiovascular diseases among non-cardiomyopathy patients, n (% of non-cardiomyopathy patients)** | Heart failure | 437 (1.1%) | 1,429 (53.5%) | 57,657 (3.7%) | 5,964 (3.8%) | 131 (16.0%) | NA | | |
| | Ischemic heart disease | 2,933 (7.1%) | 1,482 (55.5%) | 129,414 (8.3%) | 10,834 (6.9%) | 419 (51.0%) | | | |
| | Pulmonary heart/vascular disease | 53 (0.1%) | 197 (7.4%) | 17,705 (1.1%) | 1,571 (1.0%) | 19 (2.3%) | | | |
| | Pericardial disease | 152 (0.4%) | 133 (5.0%) | 12,394 (0.8%) | 740 (0.5%) | 26 (3.2%) | | | |
| | Arrhythmia/conduction disease | 2,614 (6.4%) | 1,390 (52.1%) | 179,175 (11.6%) | 10,976 (6.9%) | 285 (34.7%) | | | |
| | Hypertension | 8,429 (20.5%) | 1,017 (38.1%) | 314,882 (20.3%) | 29,887 (18.9%) | 280 (34.1%) | | | |
| | Myocarditis | 3 (<0.1%) | 198 (7.4%) | 563 (<0.1%) | 39 (<0.1%) | 123 (15.0%) | | | |

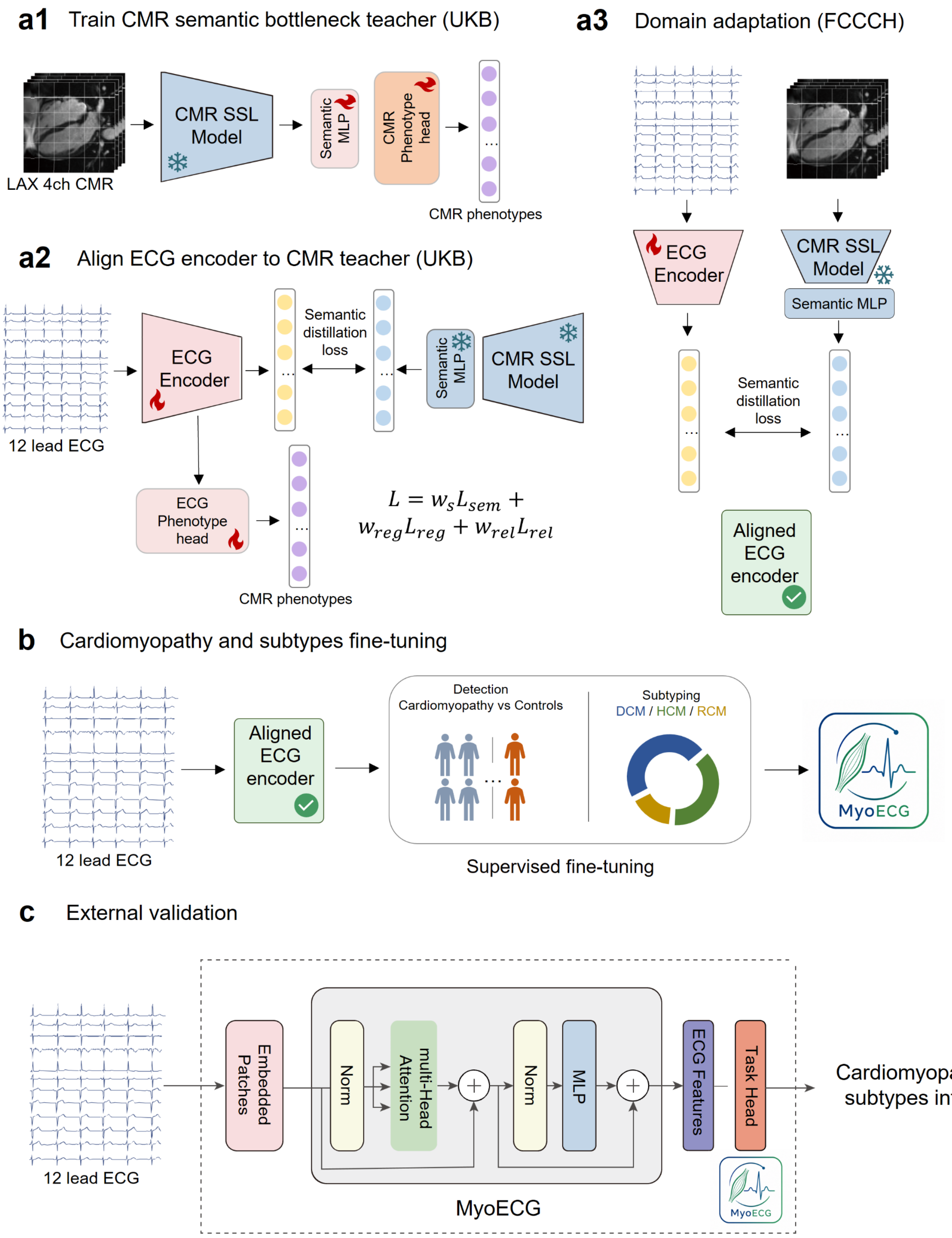

**Fig. 2 | ECG–CMR alignment for cardiomyopathy detection.** **a**, Three-step ECG–CMR pretraining. a1, A CMR semantic bottleneck teacher was first constructed in UKB using long-axis four-chamber cine CMR. A pretrained self-supervised CMR encoder extracted CMR representations, which were passed through a semantic multilayer perceptron and a phenotype head to predict standardized CMR-derived structural and functional phenotypes. a2, The ECG encoder was then aligned to the fixed CMR teacher in UKB. Paired 12-lead ECG and CMR data were used to optimize the ECG encoder through semantic distillation, ECG-to-CMR phenotypes regression and phenotype-guided relational alignment. a3, The ECG–CMR semantic alignment was further adapted in FCCCH using paired ECG–CMR data from a disease-enriched clinical distribution. During this step, the CMR teacher remained fixed and the ECG encoder was updated using semantic distillation, yielding the final aligned ECG encoder. **b**, Supervised fine-tuning. After ECG–CMR

pretraining, the CMR branch was discarded and the aligned ECG encoder was fine-tuned on FCCCH ECGs and cardiomyopathy labels for binary cardiomyopathy detection and subtype classification of dilated cardiomyopathy, hypertrophic cardiomyopathy and restrictive cardiomyopathy. The resulting ECG-only model was termed MyoECG. **c**, ECG-only inference and external validation. During validation and deployment, MyoECG required only a routine 12-lead ECG. ECG waveforms were converted into embedded patches and processed by the aligned ECG encoder and task-specific head to generate predictions for cardiomyopathy detection and subtype inference. Flame icons indicate trainable modules, and snowflake icons indicate frozen modules.

# Methods

## MyoECG dataset curation

An overview of the study design is provided in Fig. 1, and the characteristics of the study resources are summarized in Table 1 and Supplementary Tables 1–8. This retrospective study included eight datasets spanning seven clinical and research data sources, comprising 1,801,712 ECGs. Detailed eligibility criteria, disease-label ascertainment, temporal linkage, patient-level ECG selection and participant flow are provided in the Supplementary Methods and Supplementary Figs. 1–8.

Multimodal pretraining used 45,241 paired ECG–CMR examinations, including 41,128 participants from the UK Biobank (UKB)[19] and 4,113 patients from Fuwai Central China Cardiovascular Hospital (FCCCH). The FCCCH cohort also served as the supervised model-development cohort and included 1,444 patients with cardiomyopathy and 2,669 patients without cardiomyopathy. The subtype-labelled development subset comprised 726 patients with DCM, 431 with HCM and 47 with RCM (Extended Data Fig. 1).

External evaluation of the binary cardiomyopathy model included 1,756,068 ECGs from six independent resources: the Harvard–Emory ECG Database[20] (HEEDB; n=1,588,683), MIMIC-IV-ECG[21,22] (n=159,512), Sir Run Run Shaw Hospital (SRRSH; n=1,594), the First Affiliated Hospital of Zhejiang University (FAHZU; n=5,118) and the Second Affiliated Hospital of Zhejiang University (SAHZU; n=464). External evaluation of the three-class model included 13,282 subtype-labelled cardiomyopathy ECGs from HEEDB, MIMIC-IV-ECG, SRRSH and FAHZU. Because UKB contributed to multimodal pretraining, its 41,128 participants were analyzed separately for population-cohort evaluation and biological interpretation rather than as an independent external validation cohort.

Clinical decision-support performance was evaluated in the independent FCCCH-RS dataset, which included 403 ECGs from 260 patients with cardiomyopathy and 143 controls. Six cardiologists participated in the binary reader study, and a subset of 100 cardiomyopathy cases was evaluated by four cardiologists in the subtype reader study.

The UK Biobank resource was used under application number 681047. The MIMIC-IV-ECG and HEEDB datasets are publicly available controlled-access, de-identified datasets, and their secondary use in this study required no additional patient-level consent. The hospital datasets were approved by the Ethics Committee of Fuwai Central China Cardiovascular Hospital for FCCCH and FCCCH-RS (approval no. 2025 Ethics Review No. 99), the Clinical Research Ethics Committee of the First Affiliated Hospital, Zhejiang University School of Medicine for FAHZU (approval no. 2024 Research Ethics Review No. 0986, expedited review), the Human Research Ethics Committee of the Second Affiliated Hospital, Zhejiang University School of Medicine for SAHZU (approval no. 2025 Research Ethics Review No. 1661), and the Ethics Committee of Sir Run Run Shaw Hospital, Zhejiang University School of Medicine for SRRSH (approval no. 2025 Research Ethics Review No. 1282). All UKB participants provided written informed consent. For the retrospective hospital cohorts, the requirement for additional informed consent was waived by the respective ethics committees.

## Data processing

All ECG inputs were harmonized to10-s, 12-lead ECG recordings. ECG preprocessing and augmentation followed prior work[23]. Briefly, ECGs were resampled to 250 Hz, filtered with a band-pass filter of 0.67–40 Hz to suppress baseline wander and high-frequency noise, and standardized before model input. During training, random temporal cropping was applied as data augmentation. During validation and testing, fixed multi-crop evaluation was used for stable inference, and crop-level predictions were averaged to obtain the ECG-level prediction.

For CMR, we used the long-axis four-chamber cine CMR sequence. Each CMR sample was represented as a three-dimensional spatiotemporal tensor with height, width and temporal dimensions. A segmentation model was first applied to delineate the cardiac chambers[24], including the ventricles and atria. Based on the segmentation mask, the region outside the heart was cropped to reduce non-cardiac background information. The cropped cine CMR sequence was then resized to a spatial resolution of 96×96, and the temporal dimension was standardized to 50 frames. Image intensities were normalized to the range [−1,1] before model input. The detailed preprocessing of CMR is provided in Supplementary Methods. Because geometric transformations may alter cardiac morphology and chamber motion, cine CMR images were not subjected to geometric or spatial augmentations. The acquisition protocol for ECG and CMR can be seen in Supplementary Table 9.

## Method for ECG-CMR pretraining

The CMR encoder was initialized from our previously developed masked self-supervised model[25] trained on real and measurement-guided synthetic four-chamber cine CMR images. Details of its prior pretraining are provided in the Supplementary Methods.

To transfer CMR-derived myocardial structural and functional information into the ECG representation, we designed a teacher–student ECG–CMR pretraining strategy. The overall pretraining procedure consisted of three steps: CMR semantic teacher construction, ECG semantic alignment in UKB, and subsequent ECG–CMR semantic adaptation in the FCCCH. During CMR semantic-teacher construction, the CMR encoder was initialized from the pretrained long-axis CMR model, frozen for the first 50 epochs, and then fine-tuned only in its last two transformer blocks and final normalization layer; during both ECG alignment stages, the trained CMR encoder, semantic head, and teacher phenotype head were frozen. The ECG student encoder used the ST-MEM architecture as the backbone, which provides a transformer-based representation learner for 12-lead ECG waveforms.

First, we constructed a CMR semantic teacher using long-axis cine CMR and quantitative CMR phenotypes (Fig. 2a1). The CMR encoder was initialized from our previously developed self-supervised CMR model. Given a CMR input $x^{CMR}$, the CMR encoder $f_{CMR}$ first extracted a CMR representation, which was then passed through a semantic multilayer perceptron $g_{sem}$ to obtain a CMR semantic embedding:

$$z_{\mathrm{CMR}}^{\mathrm{scm}} = g_{\mathrm{sem}}(f_{\mathrm{CMR}}(x^{\mathrm{CMR}})). \tag{1}$$

A phenotype regression head was attached to this semantic embedding to predict 27 normalized CMR-derived quantitative phenotypes, including ventricular volume and function measurements, myocardial mass, and segmental and global wall-thickness measurements (Supplementary Table 10). The teacher was trained using a *SmoothL1* regression loss between the predicted and measured CMR phenotypes:

$$\mathcal{L}_{\mathrm{teacher}} = SmoothL1(\hat{y}_{\mathrm{pheno}}^{\mathrm{CMR}}, y_{\mathrm{pheno}}^{\mathrm{CMR}}). \tag{2}$$

Secondly, after training the CMR teacher, we aligned the ECG encoder to the frozen CMR semantic space (Fig. 2a2). For each paired ECG–CMR sample, the ECG signal $x^{ECG}$ was passed through the ECG encoder $f_{ECG}$ to obtain an ECG embedding: $z_{\mathrm{ECG}} = f_{\mathrm{ECG}}(x^{\mathrm{ECG}})$.

The corresponding CMR was passed through the frozen CMR semantic teacher to obtain the stop-gradient CMR semantic target:

$$\tilde{z}_{\mathrm{CMR}}^{\mathrm{sem}} = \mathrm{stopgrad}[g_{\mathrm{sem}}(f_{\mathrm{CMR}}(x^{\mathrm{CMR}}))]. \tag{3}$$

The primary ECG–CMR alignment objective was a semantic distillation loss based on cosine distance:

$$\mathcal{L}_{\mathrm{sem}} = 1 - \frac{1}{B}\sum_{i=1}^{B} \cos(z_{\mathrm{ECG,i}}, \tilde{z}_{\mathrm{CMR,i}}^{\mathrm{sem}}), \tag{4}$$

where B is mini-batch. This loss encouraged the ECG encoder to produce representations close to CMR-derived semantic embeddings while keeping the CMR teacher fixed, thereby transferring CMR-derived structural and functional information into the ECG encoder without jointly updating the CMR branch.

In the UKB pretraining stage, quantitative CMR phenotypes were also available and were used to further regularize the ECG representation. An ECG phenotype prediction head was attached to $z_{\mathrm{ECG}}$ to predict the same 27 normalized CMR phenotypes, with a *SmoothL1* phenotype regression loss:

$$\mathcal{L}_{\text{reg}} = SmoothL1(\hat{y}_{\text{pheno}}^{\text{ECG}}, y_{\text{pheno}}^{\text{CMR}}). \tag{5}$$

In addition, we introduced a phenotype-guided ECG relational loss to preserve the inter-sample similarity structure defined by the CMR phenotypes. Within each mini-batch, a phenotype-based similarity distribution was computed from pairwise distances between normalized CMR phenotype vectors:

$$q_{ij} = \text{SoftMax}_j\left(-\frac{\|y_{\text{pheno},i}^{\text{CMR}} - y_{\text{pheno},j}^{\text{CMR}}\|_2}{\tau_\phi}\right). \tag{6}$$

Similarly, an ECG-embedding similarity distribution was computed from pairwise cosine similarities between ECG embeddings:

$$p_{ij}^{\text{ECG}} = \text{SoftMax}_j\left(\frac{\cos(z_{\text{ECG},i}, z_{\text{ECG},j})}{\tau_{\text{ECG}}}\right). \tag{7}$$

The ECG relational loss was then defined as the Kullback–Leibler divergence between the phenotype-induced similarity distribution and the ECG-embedding similarity distribution:

$$\mathcal{L}_{\text{relECG}} = \frac{1}{B}\sum_{i=1}^{B} \text{KL}\left(q_i \parallel p_i^{\text{ECG}}\right), \tag{8}$$

where B denotes the mini-batch size. This term encouraged ECG embeddings from individuals with similar CMR structural and functional phenotypes to be closer in the ECG representation space, while separating ECG embeddings from individuals with dissimilar CMR phenotypes.

The overall loss for the UKB ECG–CMR pretraining stage was:

$$\mathcal{L}_{\text{stage1}} = \lambda_{\text{sem}}\mathcal{L}_{\text{sem}} + \lambda_{\text{reg}}\mathcal{L}_{\text{reg}} + \lambda_{\text{rel}}\mathcal{L}_{\text{relECG}}, \tag{9}$$

where $\lambda_{\text{sem}}$, $\lambda_{\text{reg}}$, $\lambda_{\text{rel}}$ were set to 1.0, 1.0, and 0.5, respectively.

Finally, we performed a second ECG-CMR semantic alignment stage using the FCCCH ECG-CMR dataset to adapt the ECG encoder to a more disease-informative target distribution, thereby exposing it to a larger number of pathological cases (Fig. 2a3). In this stage, the frozen CMR semantic teacher was again used to generate CMR semantic targets from paired CMR images. Because quantitative CMR phenotype labels were not available, the ECG encoder was optimized only with the semantic distillation loss:

$$\mathcal{L}_{\text{stage2}} = \mathcal{L}_{\text{sem}}. \tag{10}$$

After ECG–CMR pretraining, all CMR branches and phenotype prediction heads were discarded, and the semantic-aligned ECG encoder was retained as the MyoECG backbone for downstream supervised fine-tuning and cardiomyopathy detection.

**Fine-tuning and multicentre validation**

After ECG–CMR pretraining, all CMR-related modules were removed, and separate copies of the CMR-aligned ECG encoder were fine-tuned end-to-end on FCCCH training data for binary cardiomyopathy detection (binary MyoECG) and three-class subtype classification (three-class MyoECG; Fig. 2b). The latter used only cardiomyopathy-positive samples with definitive DCM, HCM or RCM labels. Both models required only ECG inputs at inference (Fig. 2c).

The binary MyoECG model was evaluated in the FCCCH test set and externally validated in HEEDB, UKB, SRRSH, FAHZU, SAHZU and MIMIC. The three-class MyoECG model was evaluated in the FCCCH subtype test set and externally validated in subtype-labelled cardiomyopathy ECGs from HEEDB, SRRSH, FAHZU and MIMIC. The primary external validation analyses used fixed models and operating thresholds without site-specific calibration or threshold re-optimization.

Separate recalibration and threshold-adjustment analyses were performed for binary MyoECG in UKB, MIMIC and HEEDB. For each cohort, we applied intercept-only recalibration so that the mean recalibrated probability matched the observed cardiomyopathy prevalence. Cohort-specific thresholds were selected to maximize PPV under constraints on sensitivity, specificity, and the number of predicted positives. Brier scores were compared before and after recalibration, and PPV, NPV, sensitivity, and specificity were compared between the original and adjusted operating points. These analyses were exploratory because recalibration, threshold selection, and evaluation used the same observations within each cohort. Details are given in the Supplementary Methods.

MyoECG was benchmarked against ECGFounder[26], ECGFM-KED[27], ST-MEM[23], FG-CLEP[28] and MERL[29]. All comparators were reimplemented using the architectures and pretrained-weight configurations specified in their official repositories and fine-tuned end-to-end using the same training framework and FCCCH data partitions as MyoECG.

For MyoECG and all comparators, checkpoints were selected based on the highest FCCCH validation AUROC for binary detection or macro-averaged one-vs-rest AUROC for subtype classification. For primary analyses, model-specific binary operating thresholds were selected by maximizing the Youden index (sensitivity + specificity − 1) on the FCCCH validation set and were fixed for evaluation in the FCCCH test set, all external validation datasets. Binary performance was assessed using AUROC, accuracy, sensitivity, specificity, PPV, NPV, F1 score and Cohen's kappa. Decision-curve analysis assessed net benefit across clinically relevant threshold probabilities. For three-class models, the predicted subtype was the class with the highest probability, and class-specific one-vs-rest and macro-averaged metrics were reported.

**Evaluation of CMR-informed training and model interpretability**

During UKB pretraining, we separately ablated structural (n=5), functional (n=5) and wall-thickness (n=17) phenotype groups from both teacher and student regression targets and the phenotype vectors used to construct relational similarity distributions. Loss-term ablations separately omitted the ECG phenotype-regression, phenotype-guided relational-alignment or semantic-alignment loss. FCCCH semantic adaptation was retained in these UKB-stage ablations. A separate stage ablation omitted FCCCH adaptation while retaining downstream supervised fine-tuning. All variants were evaluated on the FCCCH binary cardiomyopathy-detection task using the same patient partitions.

After UKB pretraining, regression-head predictions from the fixed CMR teacher and ECG student were evaluated for all 27 CMR phenotypes against automated imaging-derived reference measurements in 4,120 UKB test participants excluded from training and model selection. Performance was assessed using mean absolute error (MAE), Pearson's r and Lin's concordance correlation coefficient (CCC), with 95% confidence intervals estimated from 5,000 participant-level bootstrap resamples preserving prediction–reference pairs.

An ECG-only control matched MyoECG in architecture and parameter count and underwent ST-MEM pretraining on the same UKB and FCCCH training ECGs, using the same computational resources and pretraining duration. Downstream training and evaluation used identical patient partitions. Additional comparisons for binary detection and subtype classification included CardioNets[30], our prior ST-MEM-based ECG–CMR framework pretrained exclusively on paired UKB data. For binary cardiomyopathy detection, MyoECG was also compared with a long-axis four-chamber cine-CMR classifier using the same FCCCH data partitions and held-out test set of 823 participants.

Before downstream fine-tuning, embeddings from the CMR-aligned ECG encoder and pretrained ST-MEM were compared using the same 10,000 cardiomyopathy and 10,000 non-cardiomyopathy ECGs from HEEDB, with corresponding analyses in additional evaluation datasets. Analyses included PCA, geometric separation metrics and cross-validated linear-probe AUROC and accuracy. Embeddings were standardized feature-wise, using training-fold estimates for linear probing.

For both binary detection and subtype classification tasks, SmoothGrad absolute-gradient saliency maps were generated from class-specific logits and aggregated across participants after beat alignment. JT, QT and Fridericia-corrected QT intervals were also compared between cardiomyopathy and non-cardiomyopathy groups in HEEDB, reporting median differences and FDR-adjusted P values. Phenotype-group definitions, metric definitions and detailed procedures are provided in the Supplementary Methods.

### Reader study

We conducted a two-stage paired reader study in the independent, temporally held-out FCCCH-RS cohort to evaluate changes in cardiologist performance with MyoECG assistance. For binary cardiomyopathy detection, 403 de-identified 12-lead ECGs, comprising 260 cardiomyopathy cases and 143 controls, were interpreted by six cardiologists, including two residents, two attending cardiologists and two senior cardiologists. Readers were blinded to the reference labels, CMR findings, other clinical information and case prevalence. Each ECG was first interpreted using ECG alone and, after a 1-month washout period, reinterpreted in a rerandomized order with access to the MyoECG cardiomyopathy probability and model-generated classification based on the prespecified FCCCH validation-set threshold. No diagnostic feedback was provided between stages.

An exploratory subtype reader study included 100 randomly selected cardiomyopathy ECGs, comprising 44 DCM, 40 HCM and 16 RCM cases, and was performed by the four more experienced cardiologists. Using the same paired design and 1-month washout period, readers classified each ECG as DCM, HCM or RCM first using ECG alone and then with access to the MyoECG subtype probabilities and model-generated subtype classification.

Reader and standalone MyoECG performance were evaluated against the reference clinical labels. Stage effects on case-level correctness were assessed using crossed reader–case mixed-effects models, whereas reader-level paired changes in aggregate performance metrics were evaluated as complementary analyses. Detailed reader characteristics, endpoint definitions, paired tests, effect-size estimates and sensitivity analyses are provided in the Supplementary Methods.

### MyoECG score calculation

MyoECG scores were generated using five-fold cross-fitting in UKB. For each fold, the ECG–CMR representation model was pretrained using participants from the other four folds, subsequently fine-tuned on the FCCCH training set, and applied to the held-out UKB fold to predict cardiomyopathy probabilities. Predictions were pooled across the five folds and transformed using a rank-based inverse-normal transformation to obtain continuous MyoECG scores:

$$s_i = \Phi^{-1}\left(\frac{\text{rank}(p_i) - 0.5}{n}\right), \tag{11}$$

where $\Phi^{-1}(\cdot)$ denotes the inverse cumulative distribution function of the standard normal distribution and n is the number of individuals. The transformed value $s_i$ was used as the MyoECG score in genetic, proteomic and phenome-wide analyses.

Importantly, the MyoECG score was not intended as a clinical risk score. Rather, it served as a continuous analytical representation of the model output for biological characterization, enabling us to assess whether the representations learned by MyoECG were associated with established genetic and molecular correlates of cardiomyopathy.

**Genetic, phenotypic and proteomic analyses**

A genome-wide association study of the MyoECG score was performed in 41,128 UKB participants with available genotype data using PLINK (version 1.90 beta, https://www.cog-genomics.org/plink/)[31] under an additive genetic model adjusted for age, sex and the first ten genetic principal components. Genome-wide significance was defined as $P<5\times10^{-8}$, and associated loci were identified and annotated using FUMA (https://fuma.ctglab.nl/)[32]. SNP-based heritability and genetic correlations with overall cardiomyopathy and cardiomyopathy subtypes were estimated using linkage disequilibrium score regression (LDSC, https://github.com/bulik/ldsc)[33]. External summary statistics for overall cardiomyopathy, HCM and RCM were obtained from FinnGen (https://r12.finngen.fi/)[34], whereas DCM summary statistics were obtained from the HERMES Consortium (https://www.hermesconsortium.org)[35].

Bayesian colocalization analyses were performed using the R package "coloc" (https://chr1swallace.github.io/coloc/index.html)[36,37] within 500-kb windows centred on MyoECG-score GWAS lead variants. Colocalization between the MyoECG score and cardiomyopathy was assessed using the posterior probability of a shared causal variant (PP.H4), with PP.H4 ≥ 0.80 considered supportive evidence. Bidirectional two-sample Mendelian randomization analyses were performed using the R package "TwoSampleMR" (https://mrcieu.github.io/TwoSampleMR)[38] to evaluate potential causal relationships between genetic liability to cardiomyopathy and the MyoECG score, with the inverse-variance-weighted method used as the primary analysis. Corresponding genetic-correlation and MR analyses were performed for the ST-MEM baseline score. Correlations of the MyoECG and ST-MEM scores with 82 measured cardiac imaging phenotypes[39] were also compared.

Participant-level phenome-wide association analysis (PheWAS) was performed in 41,128 UKB participants with available scores and complete covariate data. Hospital-recorded ICD-10 diagnoses were mapped to phecodes using the WHO ICD-10–phecode v1.2 beta mapping and official parent-phecode aggregation rules. Cases had at least one corresponding diagnosis, and controls had no recorded diagnosis of the phenotype. Phenotypes with at least 100 cases and 100 controls were included, yielding 552 phecodes. Associations were assessed using logistic regression adjusted for age at imaging assessment and sex; sex-specific phenotypes were analysed within the relevant sex with age adjustment alone. Odds ratios and 95% confidence intervals were reported per 1-SD increase in the MyoECG score. Statistical significance was assessed using two-sided Wald tests with a Bonferroni threshold of $P < 0.05/552$. All available diagnosis records were used, so these analyses assessed associations with recorded diseases rather than incident disease risk.

Baseline plasma protein measurements were obtained from the UKB Pharma Proteomics Project using the Olink Explore platform[40]. Associations were tested for 1,463 proteins, with protein-specific sample sizes ranging from 1,381 to 5,258. For each protein, linear regression was performed with the MyoECG score as the outcome and protein abundance as the predictor, adjusting for age, sex, body mass index, ethnicity and the first ten genetic principal components, using participants with complete data for that analysis. Complementary proteome-wide MR used protein quantitative trait locus summary statistics from INTERVAL (https://www.intervalstudy.org.uk)[41] as exposures and the cross-fitted MyoECG-score GWAS as the outcome. Analyses required at least two valid independent genetic instruments, with inverse-variance weighting as the primary method. The MR summary comprised 2,059 protein assay entries, of which 1,152 yielded valid primary estimates. The Benjamini–Hochberg procedure controlled the false discovery rate separately across measured-protein associations and valid primary protein MR estimates. Proteome-wide MR was considered exploratory and hypothesis-generating. Detailed quality-control criteria, locus definitions, harmonization procedures, instrument selection and sensitivity analyses are provided in the Supplementary Methods.

**Statistical analysis**

Unless otherwise specified, 95% confidence intervals for individual model-performance estimates were obtained using 1,000 non-parametric bootstrap resamples. For pairwise model comparisons, differences in AUROC were assessed using DeLong's test for binary classification and separate class-specific one-vs-rest DeLong tests for multiclass classification. As complementary measures used only for pairwise model comparison, mean average precision and balanced accuracy were estimated using 1000 paired bootstrap resamples and assessed using two-sided paired permutation tests with 1000 permutations. Corresponding 95% confidence intervals were derived from the paired bootstrap distributions, and permutation P values were calculated using an add-one correction. These complementary measures were not included in the routine model-performance summaries. All statistical tests were two-sided unless otherwise specified. For genetic, proteomic and phenome-wide analyses involving multiple parallel tests, P values were adjusted using the Benjamini–Hochberg false-discovery-rate procedure, and FDR-adjusted q values were reported where appropriate.

Model-evaluation and resampling analyses were performed in Python using scikit-learn and MLstatkit, and R-based statistical analyses were performed in R version 4.4.3.

# Results

## Overview of MyoECG development and validation

We evaluated MyoECG across complementary clinical, interpretability and biological settings. For clinical validation, MyoECG was assessed for both cardiomyopathy detection and classification of three major cardiomyopathy subtypes. Across these tasks, MyoECG was compared with five representative pretrained ECG models under the same downstream training and evaluation protocol. To probe where MyoECG's performance gains came from and what ECG information the model used, we performed feature-space visualization and separability analysis, saliency-based waveform analysis and controlled ablation experiments. To assess clinical utility, we performed an independent reader study in which cardiologists interpreted ECGs before and after access to MyoECG outputs. For biological interpretation, we applied MyoECG to UKB ECGs to derive a continuous MyoECG score and investigated whether this AI-derived score was biologically grounded through genome-wide genetic, proteomic and phenome-wide analyses.

This design allowed subsequent analyses to evaluate MyoECG from four complementary perspectives: clinical performance, representation interpretability, human–AI collaboration, and biological grounding.

## MyoECG generalizes across centres and clinical contexts

We first evaluated MyoECG for cardiomyopathy detection (Fig. 3). In the internal FCCCH test set, MyoECG achieved an AUROC of 0.800 (95% CI, 0.766–0.830), compared with 0.728–0.779 for five representative pretrained ECG models evaluated under the same training and evaluation protocol (Fig. 3a).

We next applied the fixed model to six independent external datasets without additional training, parameter updating or site-specific calibration (Fig. 3b). AUROCs were 0.811 (95% CI, 0.790–0.831) in SRRSH, 0.974 (95% CI, 0.970–0.978) in FAHZU, 0.958 (95% CI, 0.939–0.974) in SAHZU, 0.819 (95% CI, 0.816–0.821) in HEEDB, 0.836 (95% CI, 0.826–0.846) in MIMIC, and 0.808 (95% CI, 0.741–0.866) in UKB. MyoECG achieved the highest AUROC among the compared models in each external dataset. Additional ROC and precision–recall curves are provided in Supplementary Figs. 9 and 10. The higher AUROCs in FAHZU and SAHZU should be interpreted in the context of their case–control design (Table 1). Statistical comparisons with the five pretrained ECG models generally favoured MyoECG, with statistically significant differences in most comparisons (Supplementary Fig. 11). Detailed results for AUROC, sensitivity, specificity, accuracy, PPV, NPV, F1 score and Cohen's kappa are provided in Supplementary Table 11.

Because predictive values depend on disease prevalence, we evaluated PPV and NPV at each dataset's observed prevalence, which ranged from 0.2% in UKB to 54.7% in SAHZU. At the prespecified checkpoint threshold, PPVs were 1.64%, 4.61% and 10.99% in the low-prevalence

UKB, MIMIC and HEEDB datasets, respectively, with corresponding NPVs of 99.9%, 99.6% and 98.9%. MyoECG had the numerically highest PPV among the compared ECG models in each of these datasets. Across FCCCH, SRRSH, FAHZU and SAHZU, MyoECG showed the lowest Brier score among the compared models and positive net benefit across a broad range of threshold probabilities (Supplementary Fig. 12 and Supplementary Table 12). In exploratory analyses of UKB, MIMIC and HEEDB, intercept-only recalibration improved Brier scores in all three datasets. Subsequent PPV-prioritized threshold adjustment increased PPV from 1.64–10.99% to 2.74–15.13%, while maintaining NPVs of 98.2–99.8% (Supplementary Table 13). These findings distinguish discrimination from calibration and predictive value, supporting MyoECG as a risk-stratification and triage tool rather than a standalone diagnostic test in low-prevalence settings.

Demographic subgroup analyses showed broadly comparable AUROCs between male and female participants across most datasets. Greater variation was observed across age strata, with generally lower AUROCs in older participants, particularly those aged ≥70 years (Fig. 3c and Supplementary Table 14).

**MyoECG distinguishes cardiomyopathy subtypes**

The three-class model was fine-tuned in FCCCH to classify DCM, HCM and RCM among cardiomyopathy-positive ECGs, and subtype-specific discrimination was mainly assessed using one-vs-rest AUROC (Extended Data Fig. 2). In the internal FCCCH test set, MyoECG achieved AUROCs of 0.949 (95% CI, 0.919–0.975), 0.959 (95% CI, 0.933–0.981) and 0.914 (95% CI, 0.852–0.967) for DCM, HCM and RCM, respectively, with a macro-AUROC of 0.941 (95% CI, 0.910–0.966). Under the same downstream training and evaluation protocol, the five pretrained ECG baseline models achieved macro-AUROCs ranging from 0.858 to 0.927. MyoECG achieved the highest macro-AUROC and the highest subtype-specific AUROC for all three cardiomyopathy subtypes in the internal test set, suggesting the potential of CMR-aligned ECG representations for subtype discrimination among patients with established cardiomyopathy.

We then applied the fixed three-class model without additional training, parameter updating or site-specific calibration to subtype-labelled cardiomyopathy-positive samples from four external datasets: MIMIC, HEEDB, SRRSH and FAHZU. MyoECG achieved macro-AUROCs of 0.823 (95% CI, 0.760–0.879) in MIMIC, 0.759 (95% CI, 0.747–0.770) in HEEDB, 0.854 (95% CI, 0.801–0.899) in SRRSH and 0.912 (95% CI, 0.900–0.923) in FAHZU. Across all four external subtype-validation datasets, MyoECG consistently outperformed the best baseline model, with absolute macro-AUROC gains of 0.033, 0.046, 0.024 and 0.027, respectively.

Detailed statistical comparisons between MyoECG and baseline models are provided in Supplementary Fig. 13. MyoECG generally outperformed the baseline models, with statistically significant differences in most comparisons. Detailed AUROC, sensitivity, specificity, accuracy, PPV, NPV, F1 score and Cohen's kappa results are provided in Supplementary Tables 15-19.

**CMR-informed training supports MyoECG performance and physiologically interpretable ECG representations**

To examine how CMR-informed training contributed to MyoECG performance and whether it produced physiologically meaningful ECG representations, we combined direct comparison with a cine-CMR model, phenotype-level fidelity assessment, feature-space and saliency analyses, and controlled ablation experiments.

First, we compared MyoECG with a classifier trained on long-axis four-chamber cine CMR using the same FCCCH test set. Among 823 participants, MyoECG achieved an AUROC of 0.7996 (95% CI, 0.7656–0.8299) and an accuracy of 69.74% (95% CI, 66.34–72.66%), compared with an AUROC of 0.8201 (95% CI, 0.7881–0.8495) and an accuracy of 71.08% (95% CI, 68.16–74.00%) for the cine-CMR model (Supplementary Table 20). Thus, the performance difference between MyoECG and this cine-CMR comparator was modest in this setting.

Because cardiac phenotype learning was included as an auxiliary loss during pretraining, we next assessed the phenotype-level fidelity of the ECG student and CMR teacher after pretraining in 4,120 independent UKB test participants excluded from training and model selection. For global wall thickness, the ECG student achieved an MAE of 0.3694 mm and a CCC of 0.7837, compared with 0.2581 mm and 0.8967, respectively, for the CMR teacher. Across the ten global ventricular volume, mass and functional phenotypes, CCCs for the ECG student ranged from 0.4614 for right-ventricular ejection fraction to 0.8315 for left-ventricular mass, whereas those for the CMR teacher ranged from 0.6880 for left-ventricular ejection fraction to 0.9443 for left-ventricular mass. Detailed performance metrics for both models across all phenotypes are provided in Source Data.

We then examined whether ECG–CMR alignment altered the organization of ECG representations before supervised cardiomyopathy fine-tuning. Using a balanced random sample of 10,000 cardiomyopathy ECGs and 10,000 non-cardiomyopathy ECGs from HEEDB, we compared features extracted from the aligned ECG encoder with those from ST-MEM, the pretrained ECG backbone used by MyoECG (Fig. 4a). PCA visualization showed clearer separation between cardiomyopathy and non-cardiomyopathy samples for the aligned encoder than for ST-MEM. Quantitative feature-separability analyses using complementary cluster-separation metrics and linear-probe cross-validation further supported improved class organization after ECG–CMR alignment. Similar analyses in additional validation datasets are provided in Supplementary Fig. 14. These findings suggest that multimodal ECG–CMR pretraining reshaped the ECG embedding space toward more cardiomyopathy-relevant feature organization before downstream supervised fine-tuning.

We also performed saliency-based waveform analysis to identify ECG regions associated with MyoECG predictions (Fig. 4b). For cardiomyopathy detection, salient regions were mainly concentrated around the QRS complex and T-wave segments, rather than being diffusely distributed across the entire ECG. Subtype-specific saliency maps for DCM, HCM and RCM showed a similar predominance of QRS- and T-wave-related regions, with variation in the relative saliency distribution across leads and subtypes (Supplementary Fig. 15). To further characterize the electrical features highlighted by these saliency maps, we compared JT, QT and QTc intervals between the cardiomyopathy and control groups in HEEDB (Supplementary Fig. 16). The cardiomyopathy group showed longer QT and QTc intervals and shorter JT intervals, a pattern consistent with broader QRS complexes and clinical experience[42]. Together, these findings provide qualitative support for the use of physiologically relevant ECG regions.

To examine which components of the training strategy contributed to downstream performance, we performed ablations using the same FCCCH cardiomyopathy detection task and patient split. The complete model achieved an AUROC of 0.7996. Removing structural, wall-thickness or functional phenotype targets yielded AUROCs of 0.7902, 0.7886 and 0.7891, respectively. Removing the ECG phenotype regression, relational alignment or semantic alignment loss yielded AUROCs of 0.7984, 0.7871 and 0.7912, respectively, whereas omitting the second pretraining stage yielded an AUROC of 0.7867 (Supplementary Table 21). The complete model therefore had the highest AUROC point estimate in each comparison, with the smallest numerical reduction after removal of the regression loss and larger reductions after removal of wall-thickness supervision, relational alignment or the second pretraining stage. These ablations support contributions from multiple components of the CMR-informed training scheme.

To further distinguish CMR-informed training from ECG-only pretraining, we evaluated an ST-MEM control matched for ECG architecture, parameter count, pretraining data and computational resources. Additional ECG-only pretraining did not improve the AUROC relative to the original ST-MEM comparator (0.7777 versus 0.7792), whereas MyoECG achieved an AUROC of 0.7996 on the same FCCCH test set (Supplementary Table 22). In broader backbone-controlled and prior-method comparisons, we benchmarked against ST-MEM, the backbone of our model, alongside CardioNets, our prior ST-MEM-based ECG–CMR alignment method pretrained exclusively on UKB ECG-CMR data. MyoECG achieved higher AUROC than both ST-MEM and CardioNets across all seven binary cardiomyopathy-detection datasets, with absolute gains of 0.020–0.062 over ST-MEM and 0.007–0.030 over CardioNets (Supplementary Tables 23 and 24). For subtype classification, MyoECG exceeded ST-MEM in all five subtype-validation datasets, with macro-AUROC gains of 0.014–0.046. Compared with CardioNets, MyoECG achieved higher macro-AUROC in four of five subtype datasets, except for FAHZU, where CardioNets was marginally higher.

Together, the cine-CMR comparison, phenotype-fidelity analysis, representation analysis, saliency findings and controlled experiments support the interpretation that MyoECG's performance was associated with the combination of CMR-phenotype-guided supervision and disease-enriched ECG–CMR adaptation, rather than with the ECG backbone or additional ECG-only pretraining alone.

**MyoECG assistance improves the diagnostic performance of clinicians**

To assess whether MyoECG provides incremental information for clinical ECG interpretation, we conducted a reader study using the independent FCCCH-RS dataset (Fig. 5). The dataset included 403 ECGs, comprising 260 cardiomyopathy cases and 143 controls. Six cardiologists with varying levels of ECG interpretation experience participated in the reader study: two cardiology residents with 3 years of experience each, two attending cardiologists with 6 years of experience each, and two senior cardiologists at the associate-chief-physician level with 10 and 11 years of experience, respectively, interpreted the same ECGs in two sequential stages: first using ECGs alone and then with access to the MyoECG probability score and AI-generated diagnosis. As a standalone model at the prespecified operating point, MyoECG achieved an accuracy of 0.829 (95% CI, 0.794–0.864), exceeding the mean accuracy of physicians using ECG alone (0.646; one-sample t-test on reader-level differences, P=0.0007). Across the six readers, mean accuracy, sensitivity, specificity, PPV, NPV and F1 score all increased after MyoECG assistance. Specifically, mean sensitivity increased from 0.569 to 0.678 (Δ=0.109, 95% CI, 0.041–0.175; paired t-test P=0.009), and mean specificity increased from 0.787 to 0.843 (Δ=0.056, 95% CI, 0.004–0.108; paired t-test P=0.040). These reader-level findings were further supported by a crossed reader–case mixed-effects sensitivity analysis accounting for both reader and case effects, which showed an improvement in sensitivity (Δ=0.109, 95% CI, 0.038–0.179; P=0.011) and a directionally consistent increase in specificity (Δ=0.056, 95% CI, -0.003 to 0.115; P=0.058). Across 2,418 paired reader–case decisions, 317 initially incorrect decisions became correct and 100 initially correct decisions became incorrect, yielding a net gain of 217 correct decisions, with corrections exceeding harmful reversals for all six readers (Supplementary Fig. 17), these results demonstrate the clinical utility of MyoECG. The confusion matrices, detailed reader-level results, paired statistical analyses and mixed-effects sensitivity analysis are provided in Supplementary Fig. 18 and Supplementary Tables 25 and 26.

We further conducted an exploratory subtype-level reader study using 100 randomly selected subtype-labelled cardiomyopathy cases from FCCCH-RS, including 16 RCM, 44 DCM and 40 HCM cases. Because ECG-based subtype differentiation is clinically challenging and not routinely used as a standalone diagnostic approach, this task was restricted to the four more experienced cardiologists. With MyoECG assistance, mean physician accuracy, macro sensitivity, macro specificity, macro PPV, macro NPV and macro F1 score all increased (Extended Data Fig.3). Specifically, mean macro sensitivity increased from 0.483 to 0.576 (Δ=0.093, 95% CI, -0.043 to 0.230; paired t-test P=0.117), and macro specificity increased from 0.742 to 0.803 (Δ=0.061, 95% CI, -0.008 to 0.129; paired t-test P=0.067). In a crossed reader–case mixed-effects sensitivity analysis, MyoECG assistance showed a directionally consistent increase in overall subtype accuracy (Δ=0.135, 95% CI, -0.016 to 0.286; P=0.066). Given the limited number of readers and the small number of RCM cases, these subtype results should be interpreted as exploratory. The subtype confusion matrices, detailed reader-level results, paired statistical analyses and mixed-effects sensitivity analysis are provided in Supplementary Fig. 19 and Supplementary Tables 27 and 28. Overall, these findings suggest that MyoECG can complement physician ECG interpretation for cardiomyopathy detection and may also help prioritize cardiomyopathy subtypes from routine ECGs.

**Genetic, proteomic and phenotypic associations of the MyoECG score**

Finally, we characterized the genetic, proteomic and phenotypic correlates of the MyoECG score in UKB (Fig. 6). We generated predictions for 41,128 participants using five-fold cross-fitting, with each participant excluded from the pretraining data used to generate their own prediction, and applied a rank-based inverse-normal transformation to obtain a continuous score. Compared with the ST-MEM baseline score, the MyoECG score showed generally stronger correlations with CMR-derived structural and functional traits across 82 cardiac imaging phenotypes, particularly LV wall thickness, myocardial mass and ventricular volumes (Supplementary Fig. 20).

A genome-wide association study identified 31 genome-wide significant SNPs ($P < 5 \times 10^{-8}$), comprising six independent significant SNPs represented by four lead variants across four genomic loci (Fig. 6a, Supplementary Table 29 and Source Data). FUMA[32] mapping identified genes including *NFIA*, *KCNQ1* and genes within the 17q21.31 region, including *PLEKHM1* and *MAPT*. *KCNQ1* has an established role in cardiac repolarization[43]. The quantile–quantile plot showed modest inflation of association statistics (λGC = 1.0642, Fig. 6b). Linkage disequilibrium score regression (LDSC) estimated the SNP-based heritability of the score at 12.2% ($h^2_{SNP}$ = 0.1224, SE = 0.0136), with an intercept close to unity (1.0008, SE = 0.0065).

We next examined shared genetic architecture with overall cardiomyopathy and its major subtypes using FinnGen[34] R12 summary statistics for overall cardiomyopathy, HCM and RCM, and HERMES Consortium summary statistics for DCM (Supplementary Table 30). The MyoECG score showed positive genetic correlations with overall cardiomyopathy ($r_g$ = 0.4091, FDR-adjusted $P = 1.02 \times 10^{-6}$) and DCM ($r_g$ = 0.3712, FDR-adjusted $P = 8.00 \times 10^{-4}$), whereas correlations with HCM and RCM did not reach significance after multiple-testing correction (Fig. 6c and Supplementary Table 31). Colocalization analysis supported a shared genetic signal at rs2474370 within *NFIA* between the MyoECG score and both overall cardiomyopathy (PP.H4 = 0.9805) and DCM (PP.H4 = 0.9608). No evaluated locus showed strong evidence of colocalization with HCM or RCM at the PP.H4 ≥ 0.80 threshold (Supplementary Figs. 21 and 22 and Supplementary Table 32).

Bidirectional Mendelian randomization provided complementary evidence regarding directionality (Fig. 6d and Supplementary Table 33). Following MR-PRESSO outlier removal, greater genetic liability to overall cardiomyopathy was associated with a higher MyoECG score using 27 instruments in the inverse-variance-weighted analysis ($\beta = 0.049$; 95% CI, 0.022–0.077; $P = 3.80 \times 10^{-4}$), with a concordant weighted-median estimate ($\beta = 0.049$; $P = 0.00332$). No reverse-direction association remained significant after FDR correction.

To examine the overlap with measured cardiac phenotypes, we repeated the GWAS with additional adjustment for all 27 measured CMR traits. The genetic correlation with overall cardiomyopathy attenuated from 0.4091 to 0.1305 and was no longer statistically significant ($P = 0.1428$; Supplementary Table 34), consistent with overlap between the genetic information captured by the score and the CMR traits used during pretraining. In a descriptive comparison, the ST-MEM score also showed a positive genetic correlation with overall cardiomyopathy ($r_g = 0.2961$; $P = 3.89 \times 10^{-5}$), numerically lower than that of MyoECG. Genetic liability to cardiomyopathy was not significantly associated with the ST-MEM score in the MR analyses (inverse-variance-weighted $P = 0.188$; Supplementary Fig. 23).

At the protein level, covariate-adjusted analyses identified 69 of 1,463 circulating proteins associated with the MyoECG score at FDR < 0.05, with protein-specific sample sizes ranging from 1,381 to 5,258 (Fig. 5e and Source Data). NT-proBNP, an established marker of myocardial wall stress[44], showed the strongest association ($P = 9.20 \times 10^{-11}$). Additional associations included SELE and PTN, which have been implicated in endothelial activation and cardiomyocyte stress responses[45,46], and DSG2[47], a desmosomal protein with an established genetic role in arrhythmogenic right ventricular cardiomyopathy. Associations were also observed for ACE2, NPPB, IGFBP2, LPL and SPOCK1, linking the score to circulating proteins involved in myocardial stress, cell adhesion, vascular biology and metabolism.

Exploratory proteome-wide Mendelian randomization yielded 1,152 valid primary estimates, with 82 nominal associations ($P < 0.05$; Source Data). Genetically predicted levels of RARRES1 ($P = 6.76 \times 10^{-4}$), FYN ($P = 0.00533$) and S100A13 ($P = 0.0115$) were positively associated with the MyoECG score. FYN has been implicated in experimental cardiac oxidative-stress signaling[48], S100A13 in experimental diabetic cardiomyopathy[49], and RARRES1 in a urinary proteomic panel associated with right ventricular dysfunction in idiopathic DCM[50]. None of the protein MR associations remained significant after FDR correction.

Participant-level phenome-wide association analysis directly tested associations between the MyoECG score and 552 disease phenotypes derived from hospital-recorded diagnoses in 41,128 UKB participants (Fig. 5f and Source Data). After adjustment for age and sex, with age-only adjustment for sex-specific phenotypes, 158 phenotypes met the Bonferroni threshold ($P < 9.06 \times 10^{-5}$). Higher scores were associated with left bundle branch block (OR per 1-SD increase, 6.96; 95% CI, 5.90–8.21), congestive heart failure (OR, 2.92; 95% CI, 2.63–3.25) and atrial fibrillation and flutter (OR, 1.87; 95% CI, 1.73–2.02). Associations also included hypertension, ischemic heart disease, obesity and type 2 diabetes, indicating that the score was related to a broader spectrum of cardiovascular and metabolic disease.

Together, these analyses link the MyoECG score to cardiomyopathy-related genetic variation, circulating proteins associated with myocardial stress, and a broader spectrum of cardiovascular and metabolic phenotypes. These findings are consistent with the score capturing disease-related biological variation and provide complementary biological context for its cardiomyopathy-related predictions. However, these supplementary analyses should not be interpreted as establishing the biological mechanisms underlying model predictions.

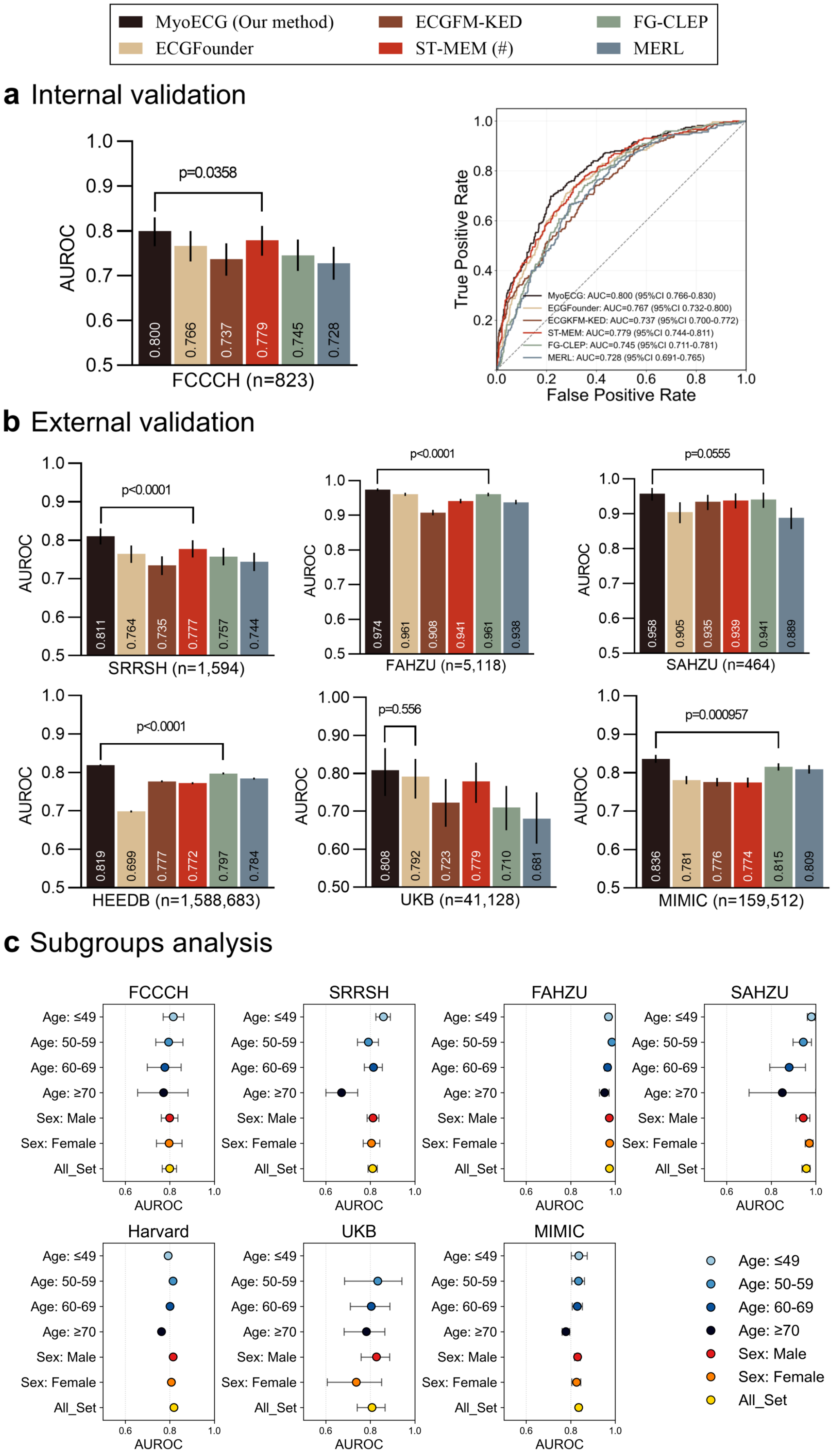

MyoECG (Our method)
ECGFM-KED
FG-CLEP
ECGFounder
ST-MEM (#)
MERL
a Internal validation
AUROC
p=0.0358
0.800
0.766
0.737
0.779
0.745
0.728
FCCCH (n=823)
True Positive Rate
False Positive Rate
MyoECG: AUC=0.800 (95%CI 0.766-0.830)
ECGFounder: AUC=0.767 (95%CI 0.732-0.800)
ECGKFM-KED: AUC=0.737 (95%CI 0.700-0.772)
ST-MEM: AUC=0.779 (95%CI 0.744-0.811)
FG-CLEP: AUC=0.745 (95%CI 0.711-0.781)
MERL: AUC=0.728 (95%CI 0.691-0.765)
b External validation
p<0.0001
0.811
0.764
0.735
0.777
0.757
0.744
SRRSH (n=1,594)
p<0.0001
0.974
0.961
0.908
0.941
0.961
0.938
FAHZU (n=5,118)
p=0.0555
0.958
0.905
0.935
0.939
0.941
0.889
SAHZU (n=464)
p<0.0001
0.819
0.699
0.777
0.772
0.797
0.784
HEEDB (n=1,588,683)
p=0.556
0.808
0.792
0.723
0.779
0.710
0.681
UKB (n=41,128)
p=0.000957
0.836
0.781
0.776
0.774
0.815
0.809
MIMIC (n=159,512)
c Subgroups analysis
FCCCH
SRRSH
FAHZU
SAHZU
Harvard
UKB
MIMIC
Age: ≤49
Age: 50-59
Age: 60-69
Age: ≥70
Sex: Male
Sex: Female
All_Set
AUROC

**Fig. 3 | Cardiomyopathy detection across internal, external datasets. a**, Internal validation. Performance of MyoECG and representative pretrained ECG models in the internal FCCCH test set. The bar plot shows AUROC values with 95% confidence intervals for MyoECG, ECGFounder, ECGFM-KED, ST-MEM (#), FG-CLEP and MERL under the same supervised fine-tuning and evaluation protocol. Significance brackets indicate comparisons between MyoECG and the best-performing baseline model, with P values calculated using DeLong's test. **b**, External validation. The fixed FCCCH-fine-tuned cardiomyopathy detection model was applied directly to independent external or public validation datasets without additional training, parameter updating or site-specific calibration. External validation datasets included SRRSH, FAHZU, SAHZU, HEEDB and MIMIC. Because UKB contributed to ECG–CMR pretraining, its performance was assessed separately as a population-cohort evaluation rather than as independent external validation. Bars show AUROC values with 95% confidence intervals for MyoECG and the five baseline ECG models. Sample sizes are indicated below each dataset name. Significance brackets indicate DeLong-test comparisons between MyoECG and the best-performing baseline model in each dataset. **c**, Subgroup analysis. AUROC values of MyoECG are shown across age and sex subgroups in FCCCH, SRRSH, FAHZU, SAHZU, HEEDB, UKB and MIMIC. Age subgroups included participants aged ≤49, 50–59, 60–69 and ≥70 years. Sex subgroups included male and female participants. The All_Set point indicates the overall AUROC in each dataset. Points represent AUROC estimates, and horizontal error bars indicate 95% confidence intervals. The # symbol indicates that ST-MEM was used as the backbone architecture for MyoECG. HEEDB, Harvard–Emory ECG Database; FCCCH, Fuwai Central China Cardiovascular Hospital; SRRSH, Sir Run Run Shaw Hospital; FAHZU, the First Affiliated Hospital of Zhejiang University School of Medicine; SAHZU, the Second Affiliated Hospital of Zhejiang University School of Medicine; MIMIC, MIMIC-IV-ECG dataset; UKB, UK Biobank.

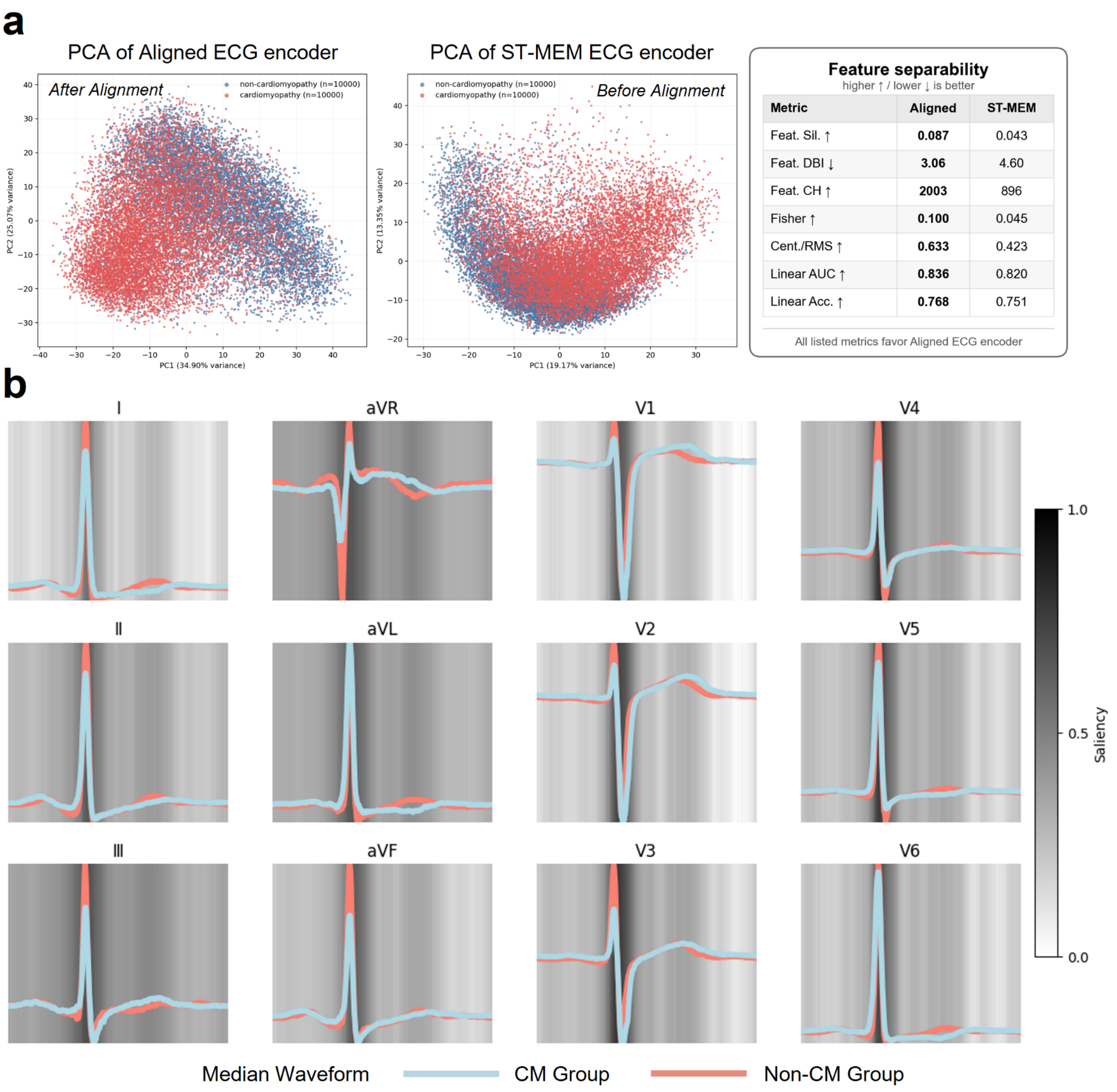

**Fig. 4 | CMR-aligned ECG representations improve feature separability and highlight physiologically relevant ECG regions. a**, Feature-space visualization and separability analysis. PCA visualization of ECG embeddings generated by the Aligned ECG encoder and a representative pretrained ECG encoder, ST-MEM, using 10,000 cardiomyopathy ECGs and 10,000 non-cardiomyopathy ECGs randomly sampled from HEEDB. The Aligned ECG encoder denotes the ECG encoder after ECG–CMR multimodal pretraining and before downstream supervised task-specific fine-tuning. ST-MEM was used as the comparator because it serves as the backbone architecture of our model, allowing the effect of ECG-CMR multimodal pretraining to be assessed relative to the underlying ECG encoder. Feature separability was quantified using complementary metrics, including silhouette score, Davies–Bouldin index, Calinski–Harabasz index, Fisher ratio, centroid-distance-to-within-class-RMS ratio, and linear-probe cross-validated AUROC and accuracy. Arrows indicate the direction corresponding to better feature separation. **b**, Saliency-based interpretability analysis for cardiomyopathy detection. Median 12-lead ECG waveforms for cardiomyopathy and non-cardiomyopathy groups are overlaid on normalized saliency maps. The background grayscale represents the relative saliency intensity from 0 to 1, with darker regions indicating ECG segments that contributed more strongly to the model output. Salient regions were mainly localized around the QRS complex and T-wave segments across multiple leads, suggesting that MyoECG relied on physiologically relevant depolarization and repolarization features rather than diffusely distributed waveform regions.

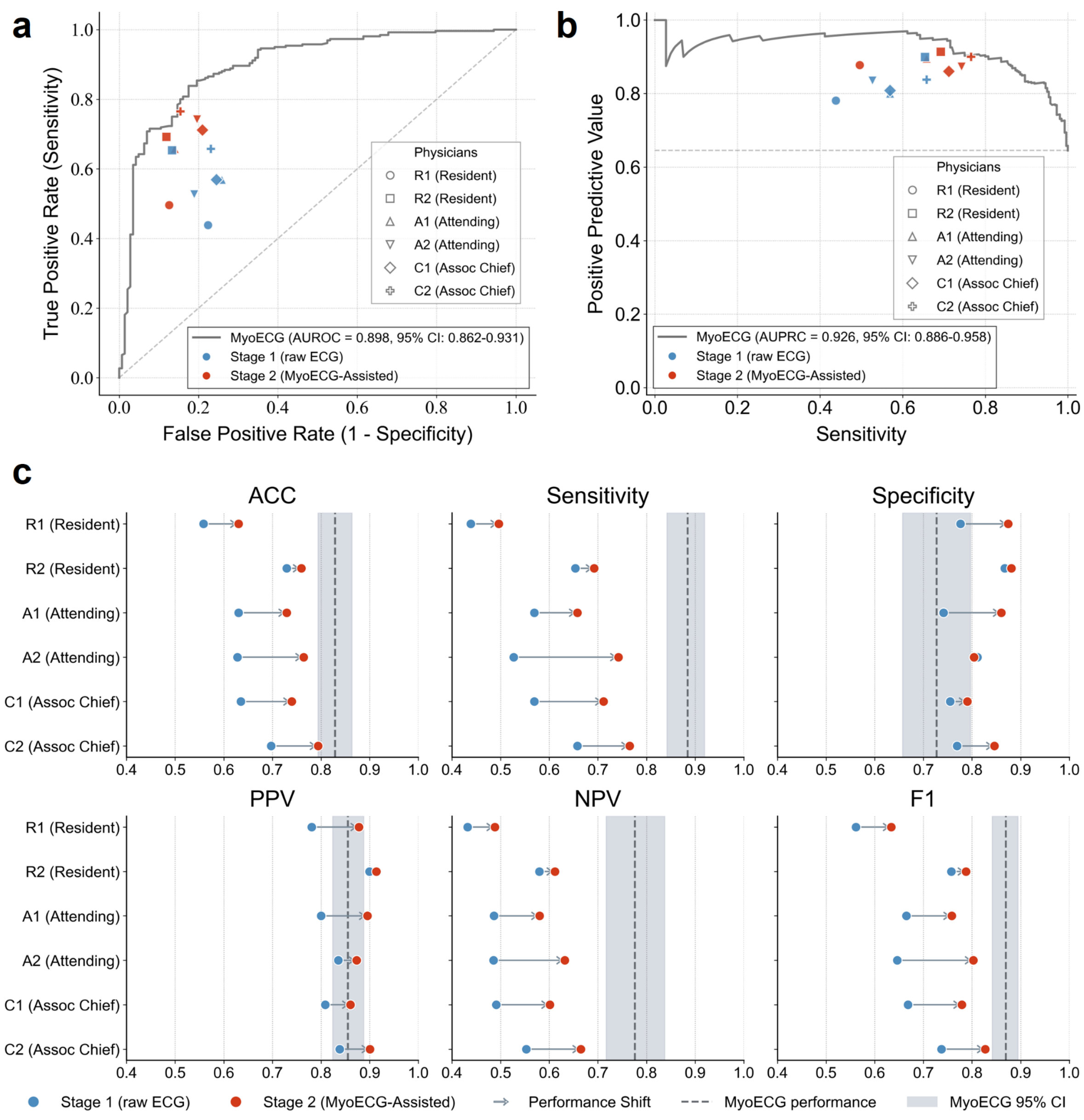

**Fig. 5 | MyoECG-assisted interpretation improves cardiologists' diagnostic performance.** A reader study was performed using the FCCCH-RS dataset, including 403 ECGs from 260 cardiomyopathy cases and 143 controls. Six cardiologists participated, including two residents, two attending physicians and two associate chief physicians. In stage 1, physicians interpreted each ECG using ECG alone. In stage 2, after a wash-out period, physicians reinterpreted the same ECGs with access to MyoECG probabilities. **a**, Receiver operating characteristic curve of the standalone MyoECG model in the reader-study subset. Individual points indicate the operating points of cardiologists before and after AI assistance. Blue points denote stage 1, and red points denote stage 2. Marker shapes indicate individual physicians. **b**, Precision–recall curve of the standalone MyoECG model and physician operating points before and after MyoECG assistance. **c**, Changes in individual physician performance across accuracy, sensitivity, specificity, positive predictive value, negative predictive value and F1 score. Arrows indicate the shift from ECG-only interpretation to MyoECG -assisted interpretation. Dashed vertical lines and shaded bands indicate the point estimate and 95% confidence interval of the standalone MyoECG model.

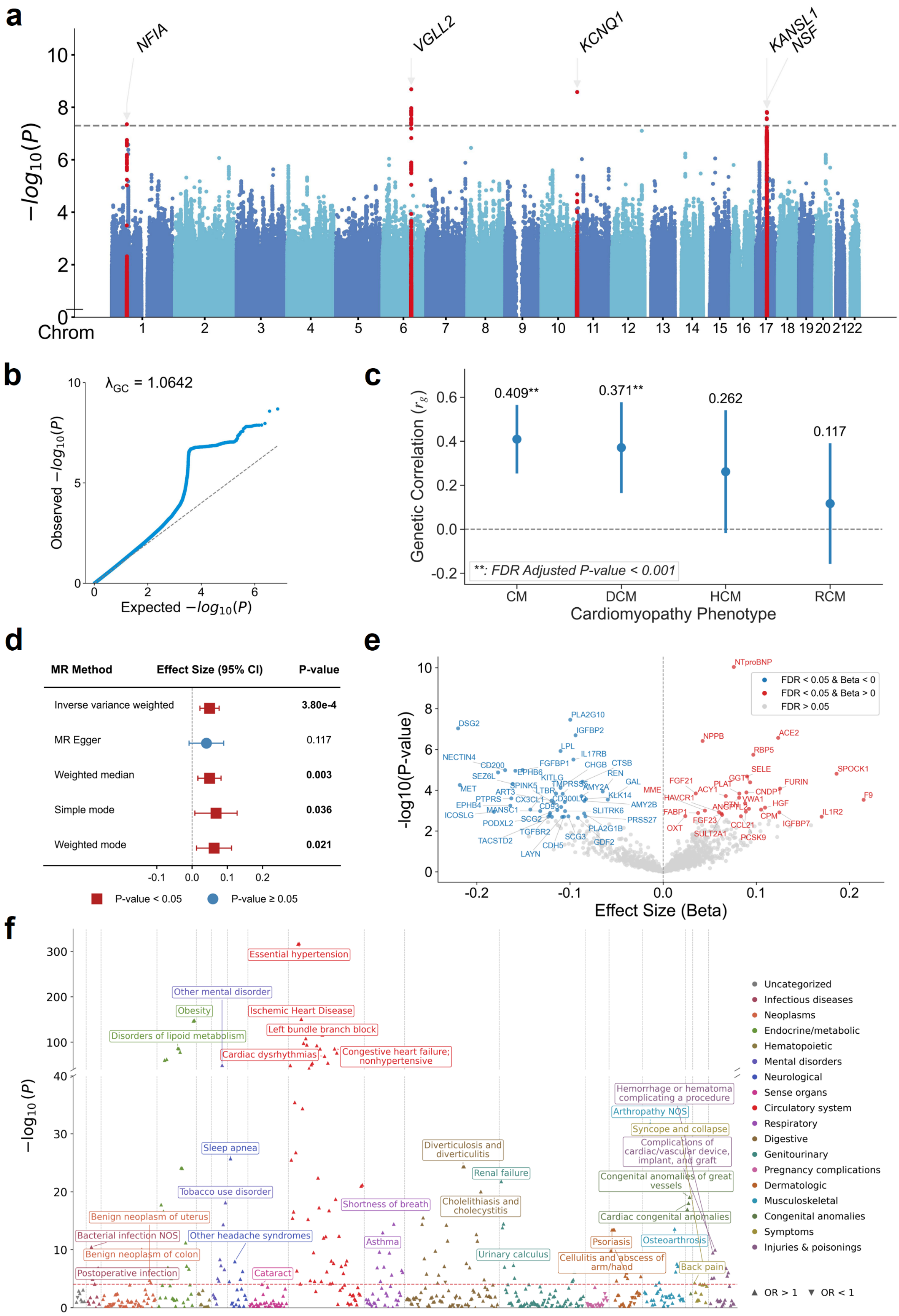

**Fig. 6 | Genetic, proteomic and phenotypic interpretation of the MyoECG score.** **a**, Manhattan plot of the genome-wide association study (GWAS) for the MyoECG score in UKB. The MyoECG score was

derived from five-fold cross-fitted cardiomyopathy probabilities in UKB using a rank-based inverse normal transformation. The horizontal dashed line denotes the genome-wide significance threshold ($P = 5 \times 10^{-8}$). **b**, Quantile–quantile plot of the MyoECG score GWAS. **c**, Genetic correlations between the MyoECG score and cardiomyopathy phenotypes estimated using linkage disequilibrium score regression (LDSC). GWAS summary statistics were obtained from FinnGen R12 for overall cardiomyopathy (CM), hypertrophic cardiomyopathy (HCM) and restrictive cardiomyopathy (RCM), and from the HERMES Consortium for dilated cardiomyopathy (DCM). Points indicate genetic-correlation estimates, and error bars indicate 95% confidence intervals. **d**, Mendelian-randomization analysis using FinnGen cardiomyopathy as the exposure and the MyoECG score as the outcome. Effect estimates and 95% confidence intervals are shown for inverse-variance weighted, MR-Egger, weighted median, simple mode and weighted mode methods. **e**, Protein association analysis. Volcano plot showing covariate-adjusted associations between baseline circulating protein levels and the MyoECG score. The x axis shows the regression effect size, and the y axis shows −log10(P value). Red and blue points indicate proteins significantly associated with the MyoECG score after FDR correction with positive and negative effect estimates, respectively; grey points indicate proteins not significant after FDR correction. **f**, Participant-level phenome-wide association analysis (PheWAS) of the cross-fitted MyoECG score with 552 diagnosis-derived phenotypes in 41,128 UKB participants. Associations were assessed using logistic regression adjusted for age and sex; sex-specific phenotypes were analyzed within the relevant sex with age adjustment alone. Each point represents a phenotype, coloured by disease category. Upward- and downward-pointing triangles indicate odds ratios greater than and less than 1, respectively, per 1-SD increase in the score. The y axis shows $-\log_{10}(P)$, and the horizontal dashed line denotes the Bonferroni significance threshold ($P = 0.05/552 \approx 9.06 \times 10^{-5}$).

## Discussion

MyoECG shows that CMR-derived myocardial structural and functional information can be transferred into the 12-lead ECG during training, improving cardiomyopathy detection and subtype classification with ECG-only inference. Critically, the transferred information is physiologically grounded: the learned representations are anchored to quantitative CMR phenotypes, and the resulting ECG score is genetically and proteomically aligned with cardiomyopathy biology. Together, they support a generalizable strategy for converting resource-intensive imaging into scalable, physiologically grounded digital biomarkers.

Mechanistically, MyoECG uses CMR as a training-only teacher[51], anchoring ECG–CMR alignment to quantitative CMR phenotypes and further adapting this alignment in a disease-enriched clinical cohort. Whereas conventional ECG-supervised or self-supervised objectives learn primarily from electrical signals and ECG-derived labels[52], and many multimodal systems require multiple modalities at deployment[53,54], MyoECG transfers information across modalities during training while preserving ECG-only inference[55,56]. Multiple complementary analyses support this design. Representation analyses demonstrated improved organization of cardiomyopathy-related features before task-specific optimization; comparative experiments showed benefits beyond both backbone pretraining and conventional cross-modal alignment; and saliency analyses suggested physiologically plausible decision patterns concentrated in depolarization and repolarization regions. Together, these findings indicate that CMR supervision enriches ECG representations before downstream task optimization, rather than merely improving classifier fitting.

MyoECG retained discriminative performance across seven validation datasets without site-specific retraining or recalibration, achieving AUROCs of 0.808–0.974 and outperforming all compared pretrained ECG models in each dataset. Between-dataset variation nevertheless highlights the influence of case mix and control definition. The exceptionally high AUROCs in FAHZU and SAHZU were obtained in disease-enriched case–control datasets with relatively healthy controls, creating a cleaner ECG contrast than in clinically heterogeneous cohorts whose non-cardiomyopathy groups included other cardiovascular diseases. The same distributional difference was apparent in calibration: curves for these two cohorts lay predominantly above the identity line, indicating that raw probabilities underestimated the observed event frequency, likely reflecting their higher sampled prevalence and more polarized predictions. Although MyoECG achieved the lowest Brier score among compared models in all four hospital datasets, absolute probability estimates should be reassessed and, where necessary, locally recalibrated in the intended deployment population. Lower AUROCs in older participants may similarly reflect reduced case–control contrast from age-related ECG changes and accumulated cardiovascular comorbidity. The lower but consistent AUROCs observed in clinically heterogeneous cohorts may therefore better approximate the discrimination expected in routine practice.

We next examined how MyoECG might operate in low-prevalence opportunistic case-finding, where PPV and downstream evaluation burden are more clinically consequential than AUROC alone. In UKB, MIMIC and HEEDB, MyoECG achieved the highest numerical PPV among the compared models, but absolute PPV at the prespecified threshold remained 1.64–10.99%. Exploratory intercept-only recalibration followed by PPV-prioritized threshold adjustment increased PPV to 2.74–15.13%. Although these absolute values remain modest, they correspond to a 6.3- to 13.7-fold enrichment over background prevalence. In the lowest-prevalence setting (UKB, 0.2%), this reduces the number of downstream imaging studies required per detected case from approximately 500 under unselected screening to approximately 37 — the quantity that ultimately determines whether an opportunistic pathway is operationally viable. Conversely, the high NPV in these datasets was strongly influenced by low prevalence and should not be interpreted as evidence that a negative result excludes cardiomyopathy. MyoECG is therefore best positioned as an opportunistic ECG-based case-finding and risk-enrichment tool that can operate alongside routine ECG acquisition or be applied to existing ECG archives to identify additional individuals who may benefit from echocardiography, CMR or specialist review. It is intended to complement CMR by helping prioritize patients for imaging and support resource allocation, not to replace CMR's diagnostic role, serve as a standalone diagnostic test, or act as a gatekeeper for clinically indicated imaging; a low score should not override symptoms, family history or other established indications for structural cardiac assessment. Because this use leverages routinely acquired ECGs with little additional acquisition burden, it may be particularly valuable in high-volume or resource-constrained healthcare systems where advanced cardiac imaging is less accessible.

The three-class model was developed and evaluated among patients with established cardiomyopathy and is intended to support subtype differentiation rather than population screening. In a hypothetical clinical scenario, its DCM, HCM and RCM probability estimates could inform specialist assessment after a positive binary screening result, or in patients with established or strongly suspected cardiomyopathy, helping prioritize CMR, genetic testing and subtype-specific diagnostic investigations.

In the reader study, access to MyoECG outputs improved cardiomyopathy detection across cardiologists with different levels of experience, with the clearest and most robust gain observed in sensitivity; the exploratory subtype reader study showed a similar direction of improvement. These findings suggest that MyoECG outputs can be incorporated into clinical judgement and improve physicians' diagnostic decisions, rather than remaining isolated model predictions.

To probe what the model has learned beyond diagnostic labels, we examined the continuous MyoECG score at the genetic, proteomic and phenotypic levels. Associations with CMR-derived structure and function, cardiomyopathy-related genetic signals and circulating proteins such as NT-proBNP are consistent with the score reflecting cardiac structural variation and myocardial stress. Notably, when the GWAS was repeated with additional adjustment for the 27 measured CMR traits, the genetic correlation with overall cardiomyopathy attenuated from 0.409 to 0.131 and was no longer statistically significant ($P = 0.143$). This is the expected consequence of the training design rather than a failure of the score: conditioning on these traits removes, at the genetic level, precisely the structural and functional information that MyoECG was trained to transfer to the ECG. The analysis therefore functions as a genetic ablation, supporting that MyoECG has indeed learned the CMR-derived phenotypic information it was designed to transfer, and that this inherited information accounts for much of the score's genetic link to cardiomyopathy.

Several limitations should be considered. First, all analyses were retrospective and PPV remained limited in low-prevalence datasets despite exploratory recalibration. These findings support feasibility but do not establish readiness for large-scale opportunistic case finding. Prospective multicentre studies should evaluate effects on clinical decisions, imaging prioritization, false-positive burden, downstream resource use and patient outcomes[57]. Second, the CMR teacher used a single four-chamber cine view with supervision from structural and functional measurements. Late gadolinium enhancement, T1/T2 or extracellular volume mapping, strain and atrial phenotyping were not used. The CMR-informed findings are therefore restricted to cine-derived structure and function and do not establish myocardial tissue characterization, including assessment of scar or fibrosis, or differentiation of ischaemic and non-ischaemic substrates. Third, the small number of RCM cases in several cohorts limits the precision of RCM-specific performance estimates. Findings from cohorts with fewer than 20 RCM cases should be interpreted descriptively, and further validation in larger cohorts with adequate RCM representation is needed. Fourth, the biological and interpretability analyses support plausibility but do not establish mechanisms. Most omics analyses were conducted in UKB, limiting generalizability across ancestries and disease spectra. Saliency maps localized broad QRS- and T-wave regions without identifying causal ECG features or explaining individual predictions. Replication in multi-ancestry, disease-enriched and longitudinal cohorts, together with orthogonal mechanistic validation, is needed.

In conclusion, MyoECG provides proof of concept that knowledge from an information-rich modality such as CMR can be transferred during model development to enrich representations derived from a low-cost routine test such as the 12-lead ECG, while preserving ECG-only inference. Although demonstrated here using CMR and ECG, the proposed framework is not specific to cardiomyopathy. More broadly, this work suggests that cross-modal knowledge transfer may provide a general strategy for translating information-rich but resource-intensive biomedical modalities into scalable digital biomarkers suitable for routine clinical deployment.

## Data availability

The data used for cross-modal pre-training were obtained from the UK Biobank under application number 681047. UK Biobank data are available to approved researchers through its standard access procedure (https://www.ukbiobank.ac.uk/enable-your-research/register).

Data used for downstream task evaluation were obtained from the MIMIC-IV-ECG database, available through PhysioNet under credentialed access (https://physionet.org/content/mimic-iv-ecg/1.0/), and from the Harvard–Emory ECG Database (HEEDB), available at https://bdsp.io/content/heedb/5.0/.

For the data from four private hospitals, the requirement for informed consent was waived by the respective ethics committees and institutions. The de-identified data can be shared only for non-commercial academic purposes and will require a formal material transfer agreement and a data use agreement. Requests should be submitted by emailing the corresponding authors ZH at zhengxinghuang@zju.edu.cn. All requests will be evaluated based on institutional policies to determine whether the requested data are subject to intellectual property or patient privacy obligations. Generally, all such requests for access to data will be responded to within 1 month.

All GWAS summary statistics analyzed in this study are publicly available as shown in Supplementary Table 32. The GWAS data for CM, HCM and RCM were obtained from the FinnGen DF12 project (https://r12.finngen.fi/). GWAS data for DCM were obtained from the HERMES Consortium (https://www.hermesconsortium.org).

## Code availability

The source codes about this study and data analysis in this manuscript are provided at https://github.com/Yukui-1999/MyoECG.

## Funding

This work was partially supported by the National Key Research and Development Program of China (Grant No. 2024YFC3506100), the Major Science and Technology Project of Henan Province under Grant No. 241100310100, the Technical Innovation Key Project of Zhejiang Province (2024C03023, 2025C01128), the Zhejiang Provincial Medical and Health Science and Technology Program (2025HY0432), the National Natural Science Foundation of China (Grant Nos. 82272129, 82470428, 82572380, 825B2011) and the Natural Science Foundation of Zhejiang Province (LZ26H020001, LHZSZ25H090002).

**Author contributions**

Z.D. designed the methodology, developed the models, implemented the computational framework, performed the experiments, analyzed the data, prepared the figures, and wrote the manuscript. K.W. prepared the figures, compiled the Supplementary Information, and contributed to manuscript revision. Y.H. contributed to the study design, overall manuscript structure, and critical revision of the manuscript. Z.L., Y.M., and H.L. contributed to model design and methodological discussions. D.Z., X.L., and Y.W. contributed to figure preparation. J.W., Y.C., L.W., M.C., Y.Z. and Y.W. participated in the reader study. C.Z. collected the SAHZU dataset. X.C. contributed to study design. M.P., J.S., Z.T., and M.Y. contributed to study discussions and provided critical feedback on the manuscript. Y.G. collected the FCCCH dataset and provided supervision. F.Y. and Z.J. provided guidance on the genetic and proteomic analyses. X.H. collected the SRRSH dataset. F.W. and H.Z. supervised the study. T.C. and H.Y. collected the FAHZU dataset, contributed to study design, provided medical clinical knowledge guidance, and supervised the study. Z.H. conceived and led the study, supervised the project, and revised the manuscript. All authors reviewed and approved the final manuscript.

## Competing interests

The authors declare that they have no competing interests.

## Acknowledgements

The UKB resource was used under application number 681047.

## Extended Data Figs

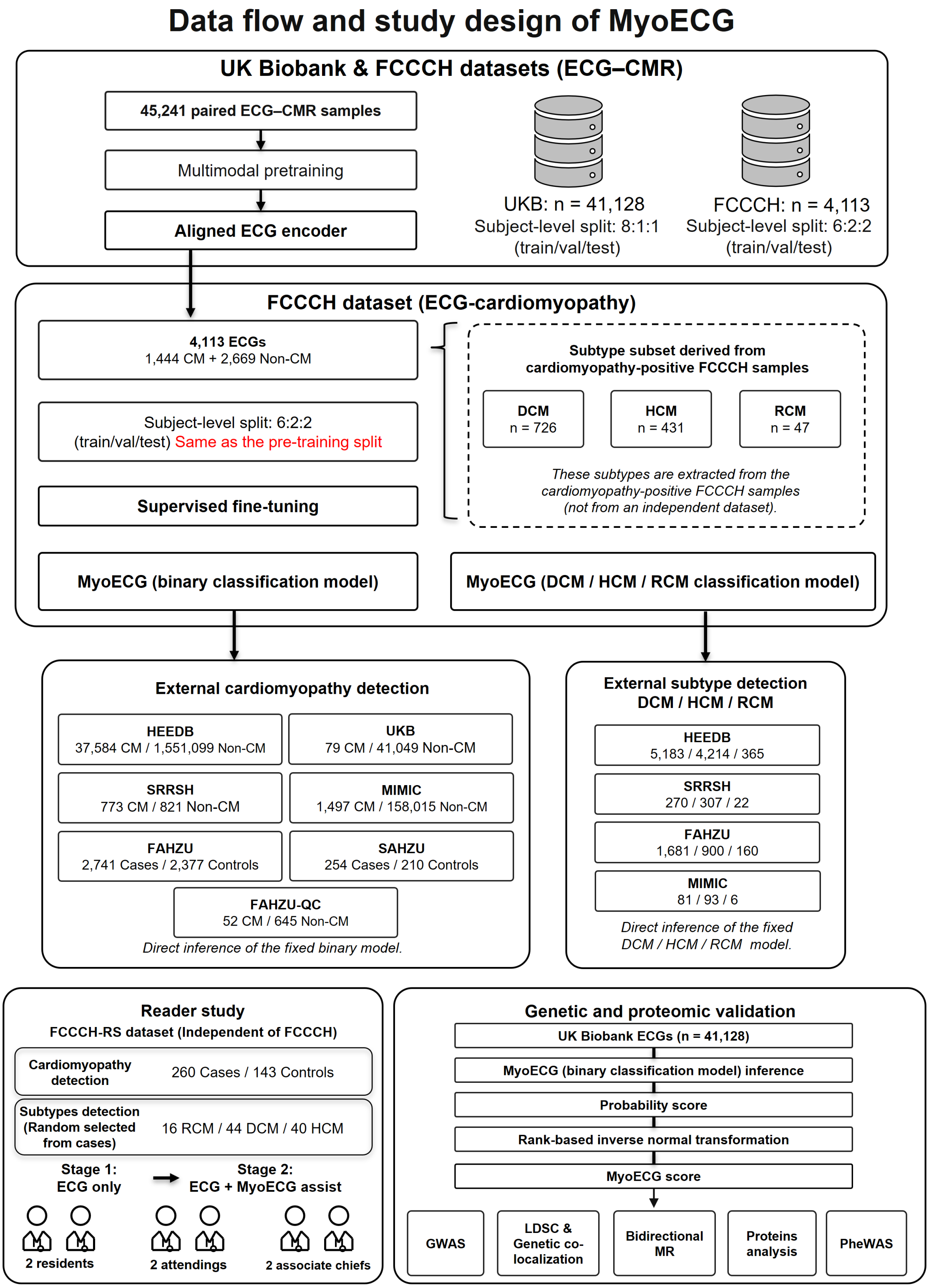

**Extended Data Fig. 1 | Data flow and study design of MyoECG.**

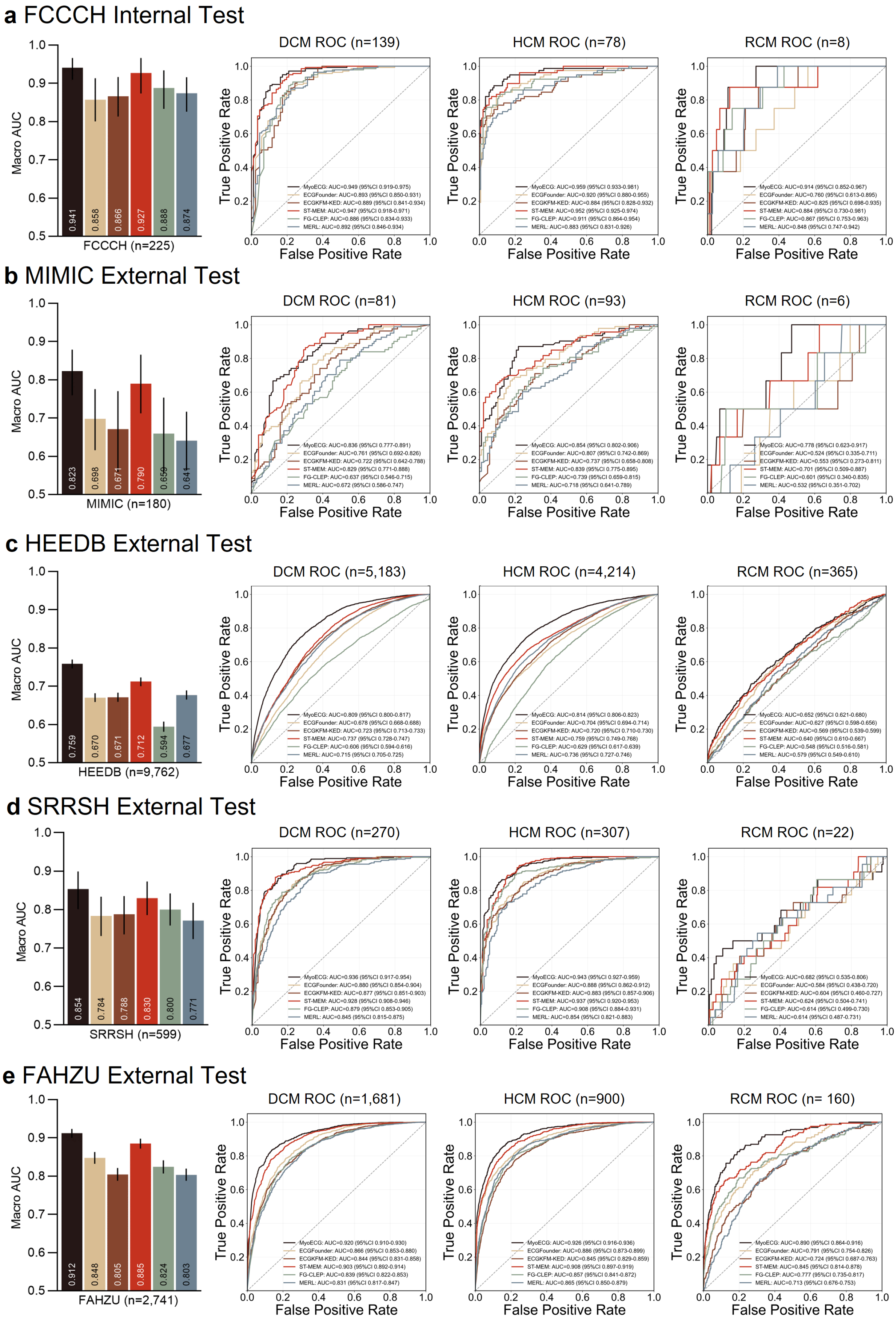
a FCCCH Internal Test
Macro AUC
0.941
0.858
0.866
0.927
0.888
0.874
FCCCH (n=225)
DCM ROC (n=139)
MyoECG: AUC=0.949 (95%CI 0.919-0.975)
ECGFounder: AUC=0.893 (95%CI 0.850-0.931)
ECGKFM-KED: AUC=0.889 (95%CI 0.841-0.934)
ST-MEM: AUC=0.947 (95%CI 0.918-0.971)
FG-CLEP: AUC=0.886 (95%CI 0.834-0.933)
MERL: AUC=0.892 (95%CI 0.846-0.934)
HCM ROC (n=78)
MyoECG: AUC=0.959 (95%CI 0.933-0.981)
ECGFounder: AUC=0.920 (95%CI 0.880-0.955)
ECGKFM-KED: AUC=0.884 (95%CI 0.828-0.932)
ST-MEM: AUC=0.952 (95%CI 0.925-0.974)
FG-CLEP: AUC=0.911 (95%CI 0.864-0.954)
MERL: AUC=0.883 (95%CI 0.831-0.926)
RCM ROC (n=8)
MyoECG: AUC=0.914 (95%CI 0.852-0.967)
ECGFounder: AUC=0.760 (95%CI 0.613-0.895)
ECGKFM-KED: AUC=0.825 (95%CI 0.698-0.935)
ST-MEM: AUC=0.884 (95%CI 0.730-0.981)
FG-CLEP: AUC=0.867 (95%CI 0.753-0.963)
MERL: AUC=0.848 (95%CI 0.747-0.942)
True Positive Rate
False Positive Rate
b MIMIC External Test
Macro AUC
0.823
0.698
0.671
0.790
0.659
0.641
MIMIC (n=180)
DCM ROC (n=81)
MyoECG: AUC=0.836 (95%CI 0.777-0.891)
ECGFounder: AUC=0.761 (95%CI 0.692-0.826)
ECGKFM-KED: AUC=0.722 (95%CI 0.642-0.788)
ST-MEM: AUC=0.829 (95%CI 0.771-0.888)
FG-CLEP: AUC=0.637 (95%CI 0.546-0.715)
MERL: AUC=0.672 (95%CI 0.586-0.747)
HCM ROC (n=93)
MyoECG: AUC=0.854 (95%CI 0.802-0.906)
ECGFounder: AUC=0.807 (95%CI 0.742-0.869)
ECGKFM-KED: AUC=0.737 (95%CI 0.658-0.808)
ST-MEM: AUC=0.839 (95%CI 0.775-0.895)
FG-CLEP: AUC=0.739 (95%CI 0.659-0.815)
MERL: AUC=0.718 (95%CI 0.641-0.789)
RCM ROC (n=6)
MyoECG: AUC=0.778 (95%CI 0.623-0.917)
ECGFounder: AUC=0.524 (95%CI 0.335-0.711)
ECGKFM-KED: AUC=0.553 (95%CI 0.273-0.811)
ST-MEM: AUC=0.701 (95%CI 0.509-0.887)
FG-CLEP: AUC=0.601 (95%CI 0.340-0.835)
MERL: AUC=0.532 (95%CI 0.351-0.702)
c HEEDB External Test
Macro AUC
0.759
0.670
0.671
0.712
0.594
0.677
HEEDB (n=9,762)
DCM ROC (n=5,183)
MyoECG: AUC=0.809 (95%CI 0.800-0.817)
ECGFounder: AUC=0.678 (95%CI 0.668-0.688)
ECGKFM-KED: AUC=0.723 (95%CI 0.713-0.733)
ST-MEM: AUC=0.737 (95%CI 0.728-0.747)
FG-CLEP: AUC=0.606 (95%CI 0.594-0.616)
MERL: AUC=0.715 (95%CI 0.705-0.725)
HCM ROC (n=4,214)
MyoECG: AUC=0.814 (95%CI 0.806-0.823)
ECGFounder: AUC=0.704 (95%CI 0.694-0.714)
ECGKFM-KED: AUC=0.720 (95%CI 0.710-0.730)
ST-MEM: AUC=0.759 (95%CI 0.749-0.768)
FG-CLEP: AUC=0.629 (95%CI 0.617-0.639)
MERL: AUC=0.736 (95%CI 0.727-0.746)
RCM ROC (n=365)
MyoECG: AUC=0.652 (95%CI 0.621-0.680)
ECGFounder: AUC=0.627 (95%CI 0.598-0.656)
ECGKFM-KED: AUC=0.569 (95%CI 0.539-0.599)
ST-MEM: AUC=0.640 (95%CI 0.610-0.667)
FG-CLEP: AUC=0.548 (95%CI 0.516-0.581)
MERL: AUC=0.579 (95%CI 0.549-0.610)
d SRRSH External Test
Macro AUC
0.854
0.784
0.788
0.830
0.800
0.771
SRRSH (n=599)
DCM ROC (n=270)
MyoECG: AUC=0.936 (95%CI 0.917-0.954)
ECGFounder: AUC=0.880 (95%CI 0.854-0.904)
ECGKFM-KED: AUC=0.877 (95%CI 0.851-0.903)
ST-MEM: AUC=0.928 (95%CI 0.908-0.946)
FG-CLEP: AUC=0.879 (95%CI 0.853-0.905)
MERL: AUC=0.845 (95%CI 0.815-0.875)
HCM ROC (n=307)
MyoECG: AUC=0.943 (95%CI 0.927-0.959)
ECGFounder: AUC=0.888 (95%CI 0.862-0.912)
ECGKFM-KED: AUC=0.883 (95%CI 0.857-0.906)
ST-MEM: AUC=0.937 (95%CI 0.920-0.953)
FG-CLEP: AUC=0.908 (95%CI 0.884-0.931)
MERL: AUC=0.854 (95%CI 0.821-0.883)
RCM ROC (n=22)
MyoECG: AUC=0.682 (95%CI 0.535-0.806)
ECGFounder: AUC=0.584 (95%CI 0.438-0.720)
ECGKFM-KED: AUC=0.604 (95%CI 0.460-0.727)
ST-MEM: AUC=0.624 (95%CI 0.504-0.741)
FG-CLEP: AUC=0.614 (95%CI 0.499-0.730)
MERL: AUC=0.614 (95%CI 0.487-0.731)
e FAHZU External Test
Macro AUC
0.912
0.848
0.805
0.885
0.824
0.803
FAHZU (n=2,741)
DCM ROC (n=1,681)
MyoECG: AUC=0.920 (95%CI 0.910-0.930)
ECGFounder: AUC=0.866 (95%CI 0.853-0.880)
ECGKFM-KED: AUC=0.844 (95%CI 0.831-0.858)
ST-MEM: AUC=0.903 (95%CI 0.892-0.914)
FG-CLEP: AUC=0.839 (95%CI 0.822-0.853)
MERL: AUC=0.831 (95%CI 0.817-0.847)
HCM ROC (n=900)
MyoECG: AUC=0.926 (95%CI 0.916-0.936)
ECGFounder: AUC=0.886 (95%CI 0.873-0.899)
ECGKFM-KED: AUC=0.845 (95%CI 0.829-0.859)
ST-MEM: AUC=0.908 (95%CI 0.897-0.919)
FG-CLEP: AUC=0.857 (95%CI 0.841-0.872)
MERL: AUC=0.865 (95%CI 0.850-0.879)
RCM ROC (n= 160)
MyoECG: AUC=0.890 (95%CI 0.864-0.916)
ECGFounder: AUC=0.791 (95%CI 0.754-0.826)
ECGKFM-KED: AUC=0.724 (95%CI 0.687-0.763)
ST-MEM: AUC=0.845 (95%CI 0.814-0.878)
FG-CLEP: AUC=0.777 (95%CI 0.735-0.817)
MERL: AUC=0.713 (95%CI 0.676-0.753)

**Extended Data Fig. 2 | Classification of cardiomyopathy subtypes across internal and external validation datasets. a,** Internal validation in FCCCH. Performance of MyoECG and representative pretrained ECG models for three-class cardiomyopathy subtype classification in the internal FCCCH test set. The bar plot shows macro-averaged one-vs-rest AUROC values with 95% confidence intervals for distinguishing dilated cardiomyopathy (DCM), hypertrophic cardiomyopathy (HCM) and restrictive cardiomyopathy (RCM). Receiver operating characteristic curves show subtype-specific one-vs-rest discrimination for DCM, HCM and RCM. **b–e,** External subtype validation. The fixed FCCCH-fine-tuned three-class model was applied directly to subtype-labelled cardiomyopathy-positive ECGs from MIMIC (b), HEEDB (c), SRRSH (d) and FAHZU (e) without additional training, parameter updating or site-specific calibration. For each dataset, the bar plot shows macro-averaged one-vs-rest AUROC values with 95% confidence intervals, and the receiver operating characteristic curves show subtype-specific discrimination for DCM, HCM and RCM. MyoECG was compared with ECGFounder, ECGFM-KED, ST-MEM, FG-CLEP and MERL under the same downstream training and evaluation protocol. Sample sizes shown below each dataset name indicate the number of subtype-labelled cardiomyopathy-positive ECGs included in the corresponding evaluation. Values in the curve legends indicate AUROC with 95% confidence intervals.

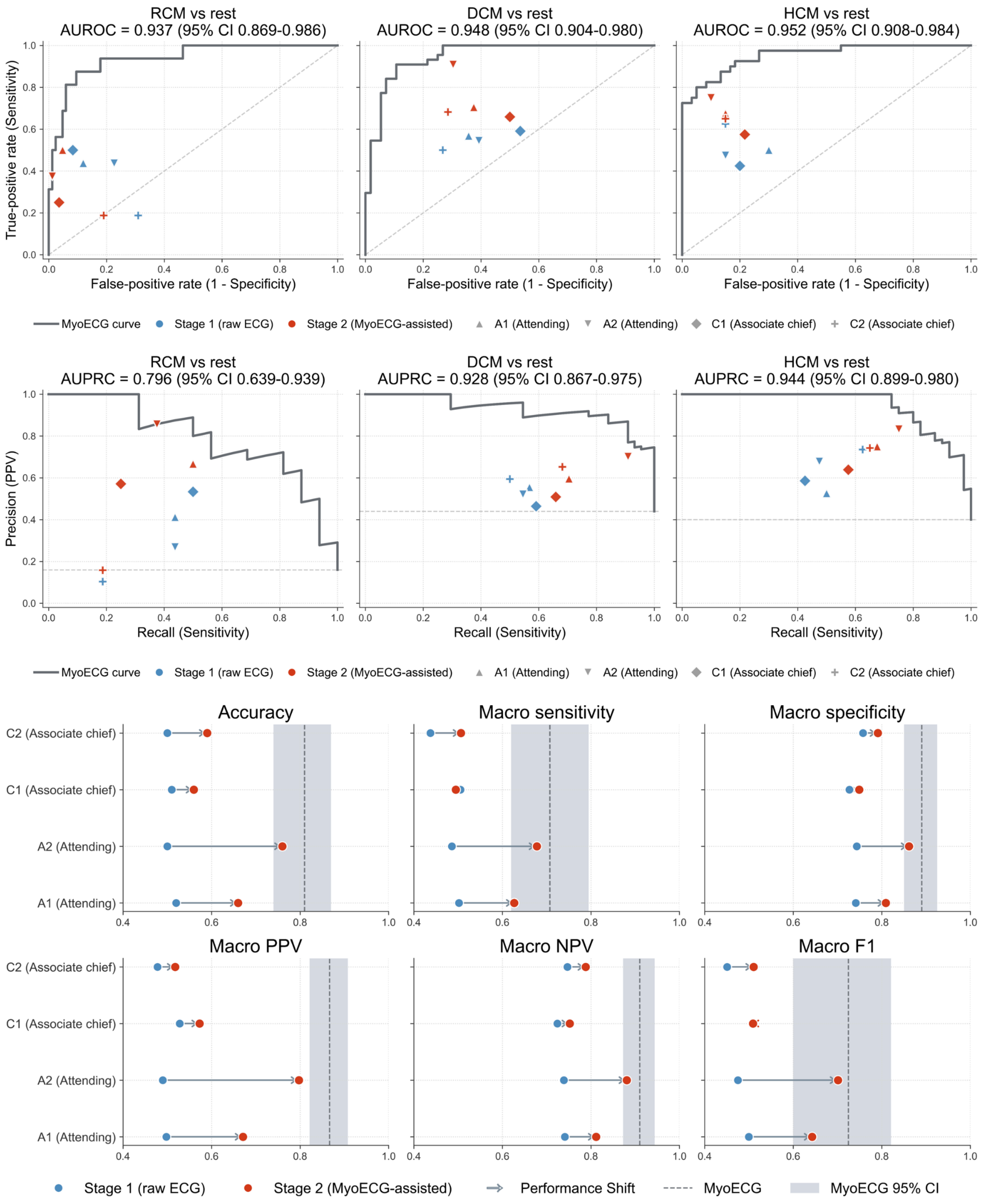

**Extended Data Fig. 3 |.** An exploratory subtype-level reader study was performed using 100 subtype-labelled cardiomyopathy ECGs randomly selected from the FCCCH RS dataset, including 44 dilated cardiomyopathy (DCM), 40 hypertrophic cardiomyopathy (HCM) and 16 restrictive cardiomyopathy (RCM) cases. Four cardiologists participated, including two attending physicians and two associate chief physicians. In stage 1, physicians classified each ECG into DCM, HCM or RCM using ECG alone. In stage 2, after a wash-out period, physicians reinterpreted the same ECGs with access to MyoECG subtype probabilities. **a**, One-vs-rest receiver operating characteristic curves of the standalone MyoECG model for RCM, DCM and HCM. Individual points indicate physician operating points before and after MyoECG assistance. Blue points denote stage 1, and red points denote stage 2. Marker shapes indicate individual physicians. **b**, One-vs-rest

precision–recall curves of the standalone MyoECG model and physician operating points before and after MyoECG assistance. **c**, Changes in individual physician performance across accuracy, macro sensitivity, macro specificity, macro positive predictive value, macro negative predictive value and macro F1 score. Arrows indicate the shift from ECG-only interpretation to MyoECG-assisted interpretation. Dashed vertical lines and shaded bands indicate the point estimate and 95% confidence interval of the standalone MyoECG model.